\documentclass[journal]{IEEEtran}
\usepackage{amssymb}
\usepackage{times}
\usepackage{graphicx}
\usepackage{subfigure}
\usepackage{multirow}
\usepackage{multicol}
\usepackage{amsmath,lipsum}
\usepackage{cuted}
\usepackage{bm}
\allowdisplaybreaks

\usepackage{dashrule}
\usepackage{algorithm}
\usepackage{algorithmic}
\usepackage[table]{xcolor}
\usepackage{threeparttable}
\usepackage{rotating}
\usepackage{dcolumn}
\usepackage{booktabs}
\usepackage{hyperref}
\usepackage{bigstrut}

\usepackage{amsthm}

\usepackage{latexsym}
\usepackage{textcomp}
\usepackage{cite}
\usepackage{colortbl}
\definecolor{light-gray}{gray}{0.75}

\usepackage[normalem]{ulem}
\newcommand\hl{\bgroup\markoverwith
	{\textcolor{yellow}{\rule[-.5ex]{2pt}{2.5ex}}}\ULon}

\usepackage{mathptmx}
\DeclareMathAlphabet{\mathcal}{OMS}{cmsy}{m}{n}   

\usepackage{epstopdf}
\usepackage{lineno}

\begin{document}

\title{Semantics-Aware Bilevel Co-Evolution: Towards Automated Multicomponent Algorithm Design }


\author{Zhiyao Zhang,
        Shenghao~Wu,
        Xingyu~Wu,
        and~Kay Chen~Tan,~\IEEEmembership{Fellow,~IEEE}
\thanks{Zhiyao Zhang, Xingyu Wu, and Kay Chen~Tan are with the Department of Data Science and Artificial Intelligence, The Hong Kong Polytechnic University, Hong Kong SAR 999077, China. (e-mail: zhiyao.zhang.cn@gmail.com; xingy.wu@polyu.edu.hk; kaychen.tan@polyu.edu.hk)}

\thanks{Shenghao~Wu is with the College of Mathematics and Informatics, South China Agricultural University, Guangzhou 510642, China. (e-mail: shwu@scau.edu.cn)}

%
}


%
%
\IEEEtitleabstractindextext{%
\begin{abstract}
Large language model (LLM)-assisted evolutionary search (LES) has emerged as a promising paradigm for automated algorithm design. However, existing LES methods usually suffer from two inherent limitations when facing the automated design of real-world complex algorithms that usually consist of multiple components. The first limitation is that they either focus on modifying entire algorithms, making it difficult to reuse high-quality components, or concentrate on component refinement within a limited set of predefined multicomponent configurations. The second limitation is the insufficient explicit modeling and exploitation of algorithm semantics (i.e., algorithm information related to in-depth algorithmic behavioral logic). These limitations severely degrade search efficiency and hinder effective exploration of complex design spaces. Therefore, this paper proposes STABLE (\emph{S}eman\emph{t}ics-\emph{A}ware \emph{B}ilevel Co-\emph{E}volution), an LES method purpose-built for automated multicomponent algorithm design that jointly introduces structural algorithm formulation and semantics-driven evolution. In STABLE, complex algorithms are organized into hierarchical and modular architectures rooted in domain knowledge, aligning the search space with their intrinsic compositional traits. Based on this structured algorithm formulation, STABLE simultaneously optimizes high-level multicomponent configurations and low-level functional components, enabling coordinated cross-level updates while maintaining suitable granularities for design space exploration. At each level, STABLE establishes a multi-faceted semantic model to assist LLMs in capturing structural correlations, functional compatibilities, and inherent rationalities among algorithm components. This semantic model serves as the core guidance for evolutionary search, enabling principled algorithm generation and informed algorithm evaluation. Extensive experiments demonstrate that the algorithms generated by STABLE outperform both human-designed baselines and those from state-of-the-art LES methods.

\end{abstract}

}

\maketitle
\IEEEdisplaynontitleabstractindextext
\IEEEpeerreviewmaketitle

\section{Introduction}\label{Sec:Int}
Optimization problems are widely encountered in numerous fields, and efficient optimization solving is vital to driving technological progress and engineering innovation~\cite{11311712}. However, the traditional design of optimization algorithms relies heavily on researchers’ professional experience and domain expertise, involving tedious processes such as operator design and parameter tuning. This not only consumes considerable time and effort but also makes it challenging to adapt to the growing complexity and scale of optimization problems. Against this background, automatic algorithm design (AAD) has become the pursuit of engineers and researchers~\cite{10993463,wu2024large,wu2025llm}.

Recently, as a phenomenal class of generative artificial intelligence, large language models (LLMs) have exhibited remarkable capabilities in natural language understanding and knowledge reasoning across diverse applications, including code generation, academic paper drafting, and scientific theorem discovery~\cite{li2026diffgraph,10767756}. This notable advancement has motivated researchers to integrate LLMs with evolutionary computation to tackle AAD tasks~\cite{liu2026systematic}. Such emerging methods are recognized as LLM-assisted evolutionary search (LES)~\cite{hu2025partition}. Distinctively empowered by population-based iterative search, LES regards each algorithm as an individual in the population and leverages LLMs as the genetic operator for algorithm generation. In this paradigm, well-designed prompts containing algorithm information (e.g. codes) serve as the core medium to guide the search direction of LLMs in exploring the algorithm space, thereby fundamentally shaping the overall performance of the generated algorithms.


To date, many LES methods have been proposed for the automated design of heuristics or single algorithm components. However, the automated design of complete multicomponent algorithms such as various multiobjective evolutionary algorithms (MOEAs) remains underexplored, which are highly relevant to practical and challenging application scenarios. For such AAD tasks, there is a fundamental mismatch between the search paradigms of existing LES methods and the structural formulation of complex evolutionary algorithms. Existing LES methods typically rely on flat formulation: treating candidate algorithms as monolithic objects for direct optimization. However, complex algorithms usually follow a hierarchical and compositional structure~\cite{zhou2011multiobjective}: high-level strategies emerge from the synergistic interaction of modular components, whose behaviors in turn depend on lower-level design configurations and algorithmic decisions. When such algorithms are constrained to a flat formulation, the search is forced to either operate on entire algorithms, resulting in coarse-grained exploration, or to optimize fine-grained components in isolation~\cite{11509330}. The former fails to reuse high-quality components during algorithm evolution, essentially rebuilding entire algorithms from scratch each generation and discarding valuable component-level knowledge. The latter, while allowing component refinement, is confined to a limited set of multicomponent configurations, unable to explore novel interactions among components beyond predefined combinations.~\cite{8454482}. As a result, the search becomes structurally ill-posed and lacks suitable granularity to effectively traverse the algorithm design space.

Beyond the  above search-formulation mismatch, a more fundamental issue persists: the search lacks an explicit characterization of the functional architecture inherent to algorithms, which belongs to intrinsic algorithm semantics, i.e., algorithm information closely tied to in-depth algorithmic behavioral logic. By ignoring component roles and their interdependencies in prompts, LLMs are restricted to superficial syntactic modifications, with little insight into how such changes influence algorithm functionality~\cite{liu2026systematic}. As a result, the search becomes semantically under-constrained. Offspring algorithms are generated by LLMs through statistical imitation of training examples, rather than through reasoning about the functional consequences of component assembly~\cite{yu2026exploring}. More critically, this shortfall propagates to the algorithm evaluation phase. Lacking semantics awareness, the search cannot reliably estimate the potential of different regions in the design space. Evaluation resources are thus allocated in a nearly uniform or weakly adaptive manner, leading to severe inefficiency. Importantly, this inefficiency stems not only from the high computational cost of algorithm performance evaluations (APEs)~\cite{yu2026exploring}, but also from the search’s inability to prioritize promising candidate algorithms according to their semantic merit.

To address the above issues, we propose \emph{S}eman\emph{t}ics-\emph{A}ware \emph{B}i\emph{l}evel Co-\emph{E}volution (STABLE), a unified framework specially tailored for automated multicomponent algorithm design (AMAD) that incorporates structured algorithm formulation and semantics-aware evolution. At its core, STABLE transcends conventional flat algorithm formulations by organizing complex algorithms as assemblies of interacting components across multiple abstraction levels based on domain-specific expert prior knowledge. This structured formulation aligns the organization of the search space with the intrinsic compositional nature of complex algorithms. Building on this formulation, STABLE establishes a bilevel co-evolutionary framework, where the upper level optimizes full algorithm structures and the lower level fauces on fine-grained modular component improvement. The framework enables coordinated cross-level evolution that systematically reuses and propagates high-quality components across emerging multicomponent configurations, transforming local design knowledge into global architectural advantage to effectively explore the complex design space. Furthermore, STABLE incorporates five‑dimensional semantic modeling to capture the inherent interdependencies among algorithm components. Operating on both code structural similarity and component functional similarity, this model lifts LLM‑driven search from the level of syntactic token rearrangement to that of semantic compositionality, enabling more semantically informed algorithm generation and algorithm performance evaluation.

The contribution of this paper is two-fold. First, we propose STABLE, which is the first LES method specially to achieve effective AMAD.  This method provides a principled and extensible framework for organizing and evolving complex algorithms in a structured manner. Second, we introduce two AMAD tasks: constrained MOEA design and surrogate-assisted MOEA design. Experiments conducted on these two tasks verify the effectiveness of STABLE.




The reminder of paper is structured in the following manner. First, Section \ref{Sec:Rew} introduces AAD, the basic workflow of LES, and related work. Next, Section \ref{Sec:Alg} offers a comprehensive description of STABLE. Subsequently, Section \ref{Sec:Set} details the experimental studies. Lastly, Section \ref{Sec:Con} concludes this paper.

\section{Preliminaries}\label{Sec:Rew}

\subsection{AAD and LES}\label{Sec:Rew1}
Without loss of generality, in current works, an AAD task is usually understood as the following abstract optimization problem:
		\begin{equation}\label{Eqn:AAD}
		\begin{aligned}
			\min : \ & F(\mathcal{A}) \\\
		\ \text{s.t.} \  & \mathcal{A} \in \Omega_{\mathcal{A}} \\
		\end{aligned}
		\end{equation}
where $\mathcal{A}$ represents the algorithm to be designed (the core decision variable of the AAD task), $\Omega_{\mathcal{A}}$ denotes the algorithm space, and $F(\cdot)$ is the evaluator used to measure the performance of $\mathcal{A}$. Specifically, $F(\cdot)$ is usually defined based on the results of $\mathcal{A}$ solving some representative optimization instances from the target optimization scenario. Obviously, (\ref{Eqn:AAD}) is black-box and lacks clear mathematical properties~\cite{liu2025fitness}.

To deal with (\ref{Eqn:AAD}), LES mainly consists of the following four steps as shown in Fig.~\ref{Fig:AAD}:

\begin{figure} [!t]
\begin{center}
    \subfigure{\includegraphics[width=1\columnwidth]{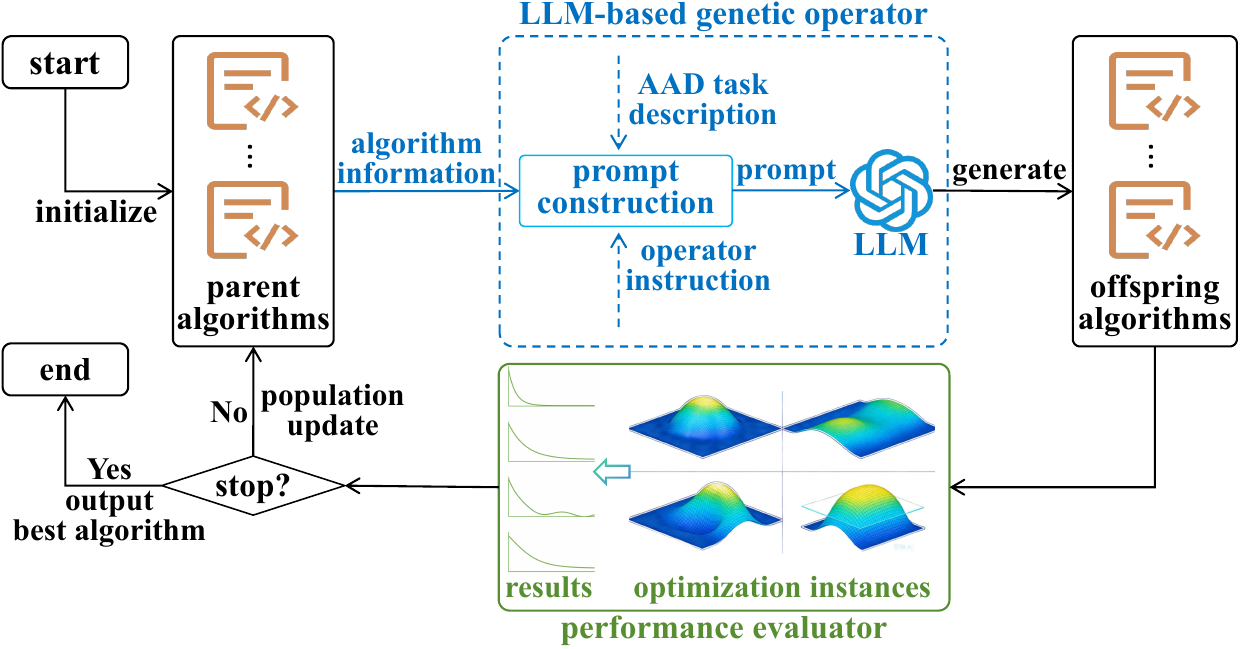}}
    \caption{Basic workflow of LES.}\label{Fig:AAD}
\end{center}
\end{figure}

\begin{enumerate}
  \item \textbf{Initialization}: A set of algorithms forms the initial population. These algorithms can be either human-designed baselines or directly generated by LLMs for the target optimization scenario.

  \item \textbf{Algorithm Generation}: Offspring algorithms are generated by the LLM-based genetic operator. Specifically, this genetic operator first constructs prompts by fusing the information of parent algorithms, the AAD task description (e.g., the characteristics of the target optimization scenario), and the operator instruction (e.g., crossover and mutation), and then uses these prompts to guide LLMs to generate offspring algorithms.

  \item \textbf{Algorithm Evaluation}: The generated offspring algorithms are used to independently solve some representative optimization instances from the target optimization scenario. The optimization results are used to define a fitness value (i.e., $F(\cdot)$ in (\ref{Eqn:AAD})) for each offspring algorithm.

  \item \textbf{Population Update}: The parent and offspring algorithms are merged into a combined pool. A ranking and selection mechanism is then applied to retain the high-quality algorithms, thereby forming an improved population for the next iteration.
\end{enumerate}
After initialization, the last three steps are repeated until the maximum number of APEs is reached, and finally the best algorithm is output. From the above introduction, algorithm generation constitutes the core of LES, where the effectiveness and design quality of prompts play a decisive role. 


\subsection{Related Work}\label{Sec:Rew3}
The rapid advancement of LLMs has provided strong momentum for AAD~\cite{wu2025towards,wu2024unlock}, leading to the emergence of LES methods. Early LES methods focus primarily on maintaining population diversity to enhance their global search capability within heuristics search spaces. For instance, FunSearch~\cite{romera2024mathematical} adopts a multi-subpopulation framework to preserve population diversity, in which subpopulations are generated in a random manner. Following FunSearch, ParEvo~\cite{hu2025partition} further introduces algorithmic features such as code similarity to guide subpopulation construction. HSEvo~\cite{dat2025hsevo} proposes two diversity evaluation metrics to assess the evolutionary progress of populations. MEoH~\cite{yao2025multi} and AutoMOAE~\cite{mo2025automoae} adopt multi-objective evolutionary frameworks to retain heuristics that satisfy diverse design requirements. As research continues to advance, increasing attention has been devoted to the positive role of algorithm semantics for prompt designs in the heuristics generation of LLM-based genetic operators. As a pioneering work, EoH~\cite{liu2024evolution} innovatively introduces algorithmic concepts to enhance LLMs’ comprehension of heuristics code. This design paradigm has been inherited and extended by subsequent studies, including EoH-S~\cite{liu2026eoh} and MoH~\cite{shigeneralizable}. ReEvo~\cite{ye2024reevo} further proposes a reflection mechanism that enables LLMs to generate interpretable hints by comparing heuristics performance, thereby accumulating high-quality heuristics design ideas. Driven by the above research progress, the core principles of the aforementioned LES methods are no longer confined to heuristics design. Instead, they have been gradually extended to broader AAD tasks. Representative applications include EvolCAF~\cite{EvolCAF} for acquisition function design in cost-aware Bayesian optimization, LLMMOP~\cite{10965770} for genetic operator design in MOEAs, and CAKE~\cite{suwandi2025adaptive} for kernel function design in Gaussian process models.

From the above review, although LES methods have achieved considerable progress, existing works mainly focus on the automated design of heuristics and individual algorithm components, with scarce attention paid to AMAD. More noteworthy is that they simply treat the entire design as a single monolithic task; that is, they regard AAD tasks as single-variable optimization problems, as formulated in (\ref{Eqn:AAD}). As analyzed in Section~\ref{Sec:Int}, when applied to AMEAD, they may suffer from inherent limitations such as search-formulation mismatch and inadequate semantic guidance. It is worth noting that although some existing LES methods have begun to enhance performance by exploiting algorithm semantics, they lack an explicit characterization of the inherent functional architecture of complex algorithms.

\section{STABLE}\label{Sec:Alg}

\begin{figure} [!t]
\begin{center}
    \subfigure{\includegraphics[width=0.55\columnwidth]{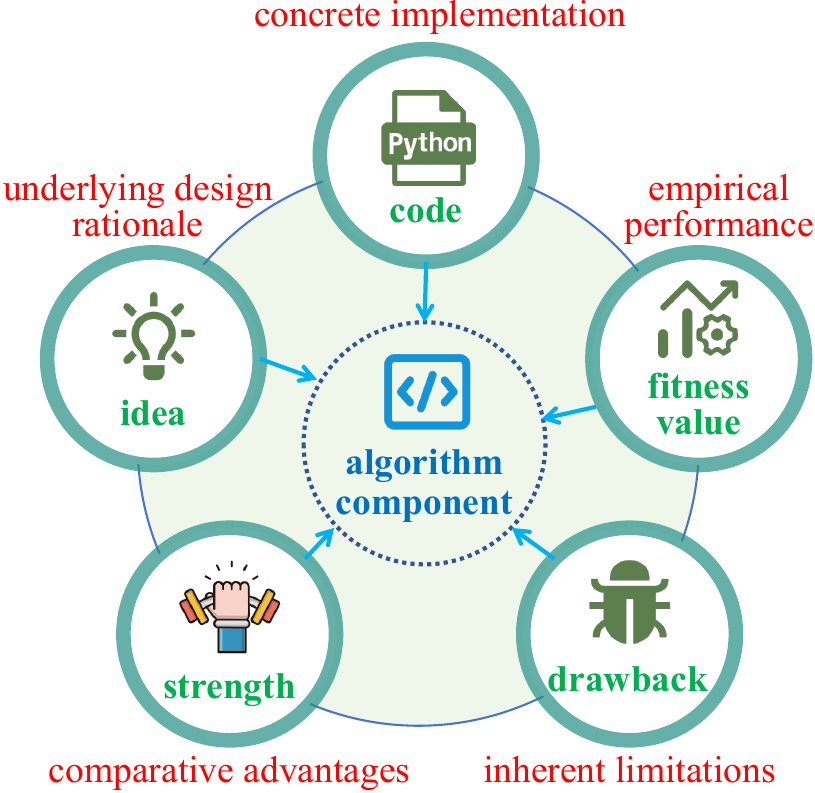}}
    \caption{Five-dimensional semantic model for each algorithm component.}\label{Fig:semantic}
\end{center}
\end{figure}

\begin{figure*} [!t]
\begin{center}
    \subfigure{\includegraphics[width=2\columnwidth]{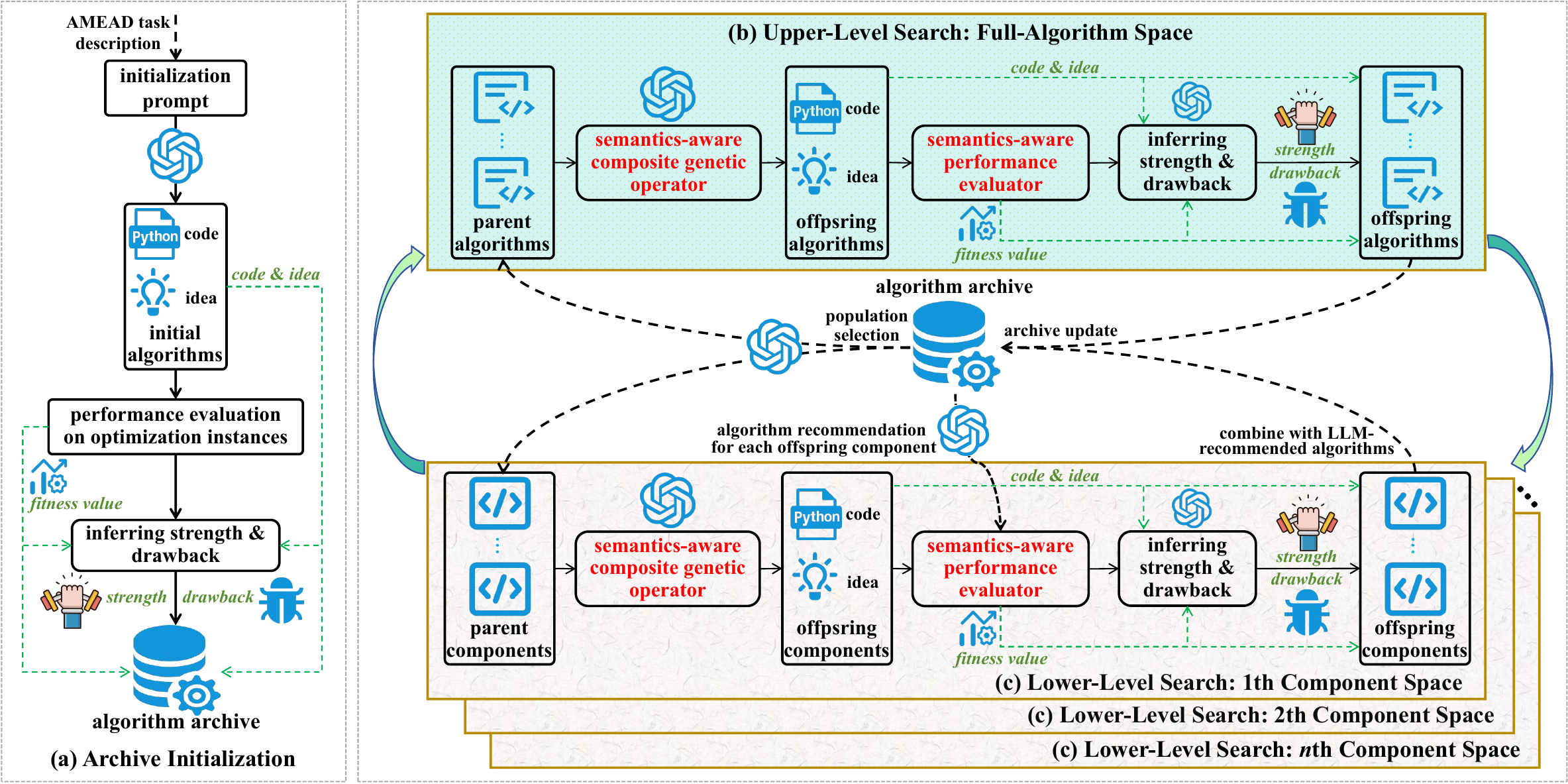}}
    \caption{Framework of STABLE. (a) Archive Initialization: An algorithm archive is initialized with $N$ complete algorithms generated by the LLM based on the initialization prompt. (b) Upper-Level Search: The full-algorithm space is explored to generate promising complete algorithms to update the algorithm archive. (c) Lower-Level Search: High-performing components are exploited in all component spaces simultaneously and are combined with potential algorithms from the algorithm archive. The bilevel search proceeds in a collaborative manner, where the \textbf{Semantics-Aware Composite Genetic Operator} and the \textbf{Semantics-Aware Performance Evaluator} are adopted to guide the algorithm/component evolution.}\label{Fig:STABLE}
\end{center}
\end{figure*}

\begin{figure} [!t]
\begin{center}
    \subfigure{\includegraphics[width=1\columnwidth]{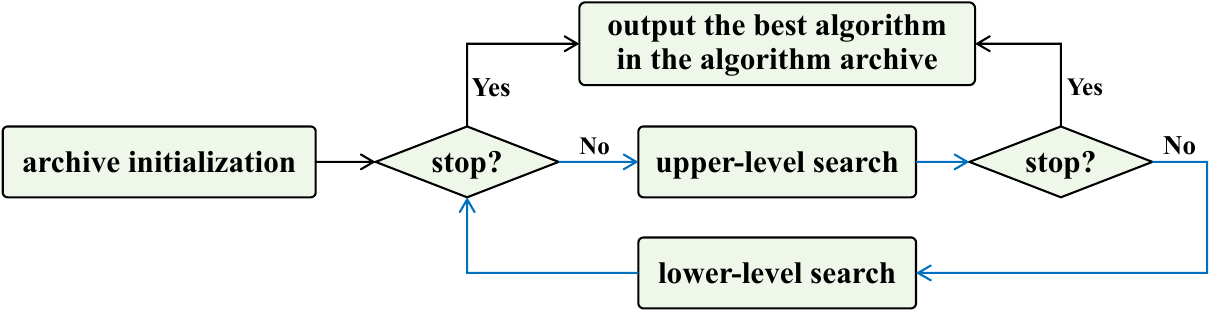}}
    \caption{Flowchart of STABLE.}\label{Fig:flowchart}
\end{center}
\end{figure}

In this study, we propose STABLE to achieve effective AMEAD, which incorporates structural algorithm formulation and semantics-aware evolution. In STABLE, an AMEAD task is explicitly formulated as follows:
\begin{equation}\label{Eqn:ACAD}
\begin{aligned}
\min : \ & F(\mathcal{A}) \\
\text{s.t.} \  & \mathcal{A}=(\mathcal{M}_1, \ldots, \mathcal{M}_n)^\top \in \Omega_{\mathcal{A}} \\
\end{aligned}
\end{equation}
where $\mathcal{A}$ is the complex algorithm consists of $n$ algorithm components derived based on domain-specific expert prior knowledge and $\mathcal{M}_i$ represents the $i$th  algorithm component. Since complex algorithms typically require numerous generations to solve the corresponding optimization problems, particularly for surrogate-assisted evolutionary algorithms addressing expensive optimization problems, $F(\cdot)$ is usually computationally highly expensive~\cite{ye2024reevo,yu2026exploring}. This inherent property severely restricts the available number of APEs.

Based on the above structural formulation, STABLE introduces a bilevel co-evolutionary framework, in which the upper level evolves over the full-algorithm space while the lower level performs searches within component spaces. The two levels operate in a collaborative and coordinated manner. To support effective search, as shown in Fig.~\ref{Fig:semantic}, each algorithm component is characterized by a five-dimensional semantic model: code, idea, strength, drawback, and fitness value. Specifically, code represents the concrete implementation; idea captures the underlying design rationale and working mechanism; strength and drawback summarize comparative advantages and inherent limitations, respectively; and fitness value quantifies empirical performance. These complementary semantics equip LLMs with rich and standardized priors to accurately capture the intrinsic properties and structural relationships of algorithm components. This deep understanding enables reliable assessment of the structural consistency, functional compatibility, and design rationality of candidate algorithms, greatly reducing invalid or semantically conflicting variations. Accordingly, such fine-grained semantics facilitate the mining and reuse of high-quality design patterns, improve the stability and interpretability of evolution, and render the entire pipeline of algorithm generation and algorithm evaluation more targeted, efficient, and consistent with domain-specific design principles.

In the next subsections, we first introduce the overall framework of STABLE and then elaborate on the two core components of STABLE: semantics-aware composite genetic operator and semantics-aware performance evaluator.

\subsection{Overall Framework}
Fig.~\ref{Fig:STABLE} depicts the overall framework of STABLE, which consists of three parts: 1) archive initialization, 2) upper-level search, and 3) lower-level search. The first part generates $N$ initial complete algorithms and initializes an algorithm archive (with a maximum capacity of $2N$) to store historically high-performance algorithms. The second part explores promising complete algorithms (multicomponent configurations) in the full-algorithm space and the third part searches for high-quality algorithm components in all component spaces simultaneously. After archive initialization, upper-level and lower-level searches are executed alternately and iteratively until the maximum number of APEs (denoted as $APE_{\max}$) is reached, and the best algorithm in the archive is finally output.

\begin{figure} [!t]
\begin{center}
    \subfigure{\includegraphics[width=0.9\columnwidth]{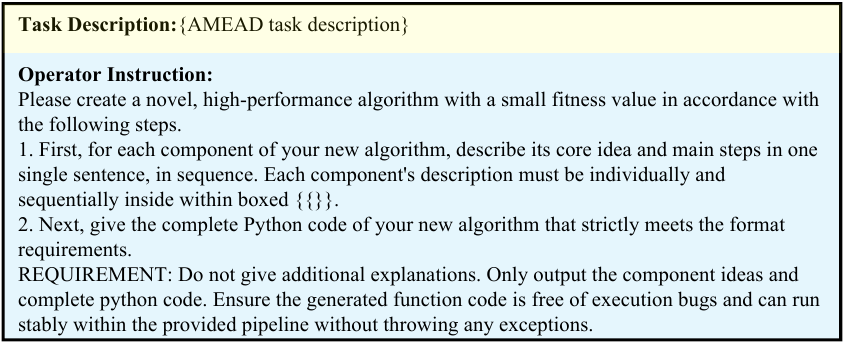}}
    \caption{Prompt for initialization.}\label{Fig:prompt_initial}
\end{center}
\end{figure}

\begin{figure} [!t]
\begin{center}
    \subfigure{\includegraphics[width=0.6\columnwidth]{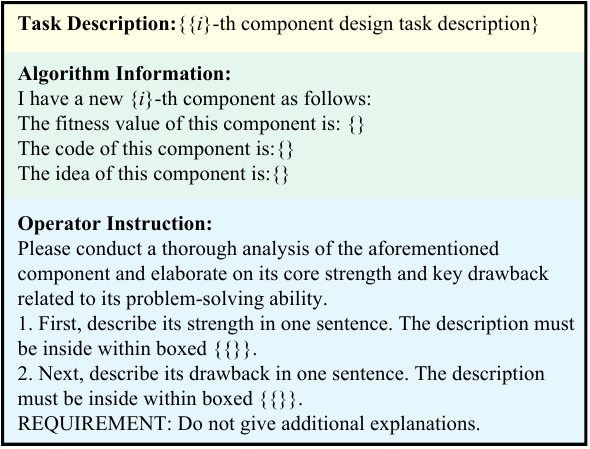}}
    \caption{Prompt for inferring strengths and drawbacks.}\label{Fig:prompt_drawback}
\end{center}
\end{figure}

\begin{figure} [!t]
\begin{center}
    \subfigure{\includegraphics[width=0.6\columnwidth]{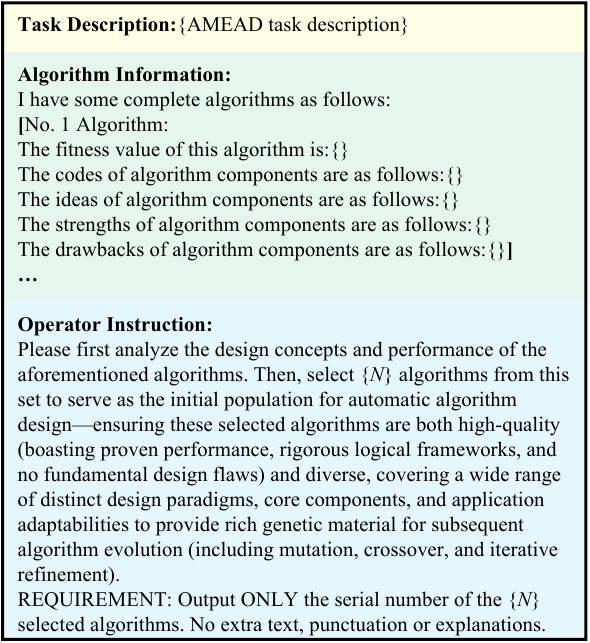}}
    \caption{Prompt for population selection.}\label{Fig:prompt_population_selection}
\end{center}
\end{figure}

\begin{figure} [!t]
\begin{center}
    \subfigure{\includegraphics[width=0.9\columnwidth]{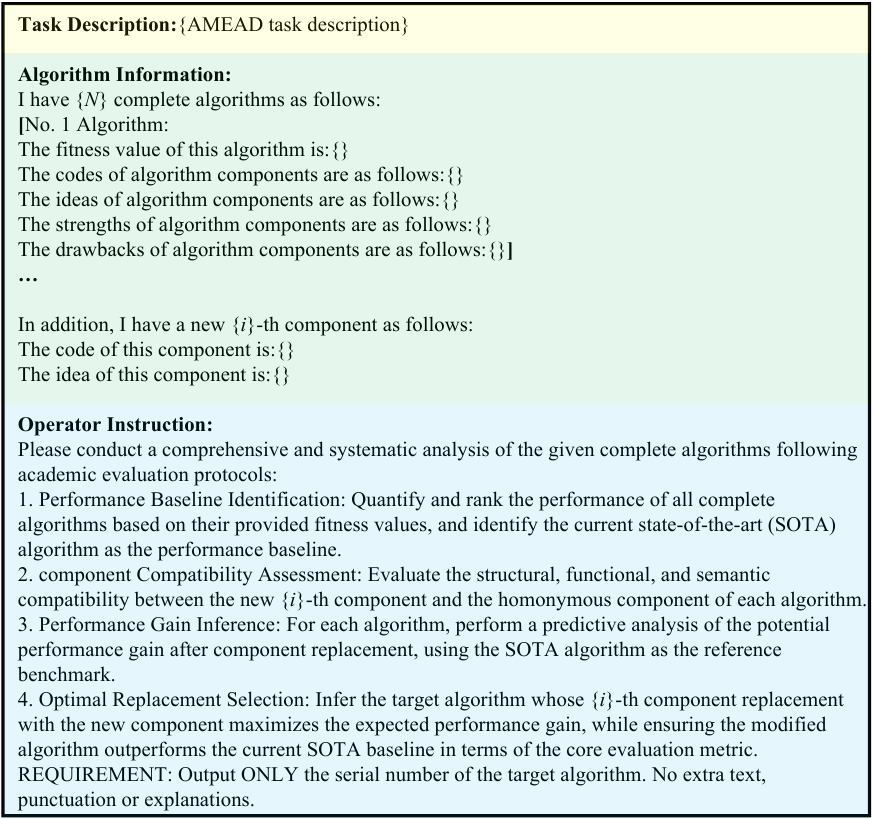}}
    \caption{Prompt for algorithm recommendation in lower-level search.}\label{Fig:prompt_algorithm_selection}
\end{center}
\end{figure}

For clarity, the detailed flowchart is shown in Fig.~\ref{Fig:flowchart}, and the detailed pseudocode of STABLE is provided in Section S-I of the supplementary material. The detailed procedures of the three parts are elaborated in the following.
\begin{enumerate}
\item \textbf{Initialization}: First, we construct an initialization prompt (as shown in Fig.~\ref{Fig:prompt_initial}) and feed it to the LLM to generate $N$ distinct complete algorithms, where each algorithm component includes code and idea. Next, these $N$ complete algorithms are evaluated on the pre-set optimization instances to obtain fitness values. Subsequently, for each component of each complete algorithm, its strength and drawback are further summarized by the LLM, as illustrated in the prompt in Fig.~\ref{Fig:prompt_drawback}. Finally, these $N$ initial complete algorithms are stored into the archive.

\item \textbf{Upper-Level Search}: First, as shown in the prompt in Fig.~\ref{Fig:prompt_population_selection}, the LLM analyzes the design concepts and performance of the algorithms in the archive, then selects $N$ high-quality and diverse ones as parent algorithms. These parent algorithms generate $N$ offspring algorithms containing codes and ideas via the semantics-aware composite genetic operator. Afterward, the semantics-aware performance evaluator is adopted to obtain the fitness value for each offspring algorithm. Then, as shown in the prompt in Fig.~\ref{Fig:prompt_drawback}, the LLM further derives the strength and drawback of each component of each offspring algorithm. Finally, these offspring algorithms are added to the archive, whose size is further truncated to $2N$ by removing inferior solutions upon overflow.


\item \textbf{Lower-Level Search}: First, consistent with upper-level search, as shown in the prompt in Fig.~\ref{Fig:prompt_population_selection}, the LLM selects $N/n$ complete algorithms from the archive; their corresponding components form the parent component population for each respective component space. Then, in each component space, the semantics-aware composite genetic operator is applied to the parent components to generate $N/n$ novel offspring components containing codes and ideas. For each offspring component, as shown in the prompt in Fig.~\ref{Fig:prompt_algorithm_selection}, its information is input into the LLM, which then recommends the most promising and compatible complete algorithm from the archive. By replacing the corresponding component in this recommended algorithm, a new complete algorithm is constructed. Finally, this new complete algorithm is evaluated via the semantics-aware performance evaluator, and is used to update the archive after the strengths and drawbacks are derived by the LLM (as shown in the prompt in Fig.~\ref{Fig:prompt_drawback}). Note that the complete algorithms recommended by the LLM for different components may vary, which enables the exploitation and reuse of diverse superior components and configurations.

\end{enumerate}

In principle,  upper-level search performs global structural exploration at the macro level by generating diverse and syntactically valid complete algorithms. This high-level search preserves population diversity within the archive, mitigates the risk of premature convergence towards local optima, and ensures the evolutionary process remains broadly oriented towards promising regions of the design space. Concurrently, lower-level search conducts fine-grained local exploitation by incrementally refining individual components within high-performing algorithm architectures preserved in the archive. This targeted refinement leverages and accumulates superior structural components, strengthens local design quality, and accelerates the convergence towards high-performance multicomponent configurations.

It should be noted that the above two search strategies are not isolated, but an organic whole that is interrelated and synergistic. Upper-level search provides a wealth of candidate algorithm architectures and potential sources of high-quality components for local exploitation, avoiding local exploitation from falling into the limitation of a single structure; the high-quality components and optimization experience accumulated by lower-level search can, in turn, feed back to global exploration, improve the basic quality of the complete algorithms generated subsequently, make global exploration more targeted, and further optimize the efficiency and effect of the entire evolutionary process.

By alternately executing these two complementary and cross-level search strategies, STABLE establishes a dynamic and adaptive balance between global exploration and local exploitation anchored on the evolving algorithm archive. Such a hierarchical co-evolutionary mechanism not only accommodates the intrinsic compositional complexity of algorithm design but also substantially boosts search efficiency and navigational effectiveness within the vast, structurally intricate design space, ultimately yielding robust improvements in optimization performance.

\begin{algorithm}[!t]
	\renewcommand{\algorithmicrequire}{\textbf{Input:}}
	\renewcommand{\algorithmicensure}{\textbf{Output:}}
	\caption{Semantics-Aware Composite Genetic Operator}\label{Alg:Operator}
	\begin{algorithmic}[1]\footnotesize
		\REQUIRE Parent population: $\mathbb{P}$
        \ENSURE Offspring population: $\mathbb{Q}$
        \STATE  $\mathbb{Q} = \varnothing$
		\STATE  \emph{k} $\leftarrow$  the optimal cluster number for $\mathbb{P}$ recommended by the LLM;
        \STATE  Compute the code structural similarity between any two individuals in $\mathbb{P}$;
        \STATE  $[\mathbb{C}_1, \ldots, \mathbb{C}_k] \leftarrow  K-Means(k, \mathbb{P})$;
        \STATE  $G_{best}$ $\leftarrow$  the best individual in $\mathbb{P}$;
        \FOR{$i = 1$ \TO $k$}
            \STATE $L_{best}$ $\leftarrow$  the best individual in $\mathbb{C}_i$;
            \FOR{each $Indiv \in \mathbb{C}_i$}
                \IF{$rand \leq 0.5$}
                    \IF{$rand \leq 0.5$}
                        \STATE {*********\emph{Intra-Cluster Crossover}**********}
                        \IF{$rand \leq 0.5$}
                            \STATE $Crossover\_Partner$ $\leftarrow$  $L_{best}$;
                        \ELSE
                            \STATE $Crossover\_Partner \leftarrow$  random individual in $\mathbb{C}_i$;
                        \ENDIF
                    \ELSE
                         \STATE{*********\emph{Inter-Cluster Crossover}**********}
                        \IF{$rand \leq 0.5$}
                             \STATE $ Crossover\_Partner$ $\leftarrow$  $G_{best}$;
                        \ELSE
                              \STATE $ Crossover\_Partner$ $\leftarrow$ random individual in $\mathbb{P}/\mathbb{C}_i$;
                        \ENDIF
                    \ENDIF
                    \STATE $Offspring$ $\leftarrow$ $Crossover(Indiv,Crossover\_Partner)$;
                \ELSE
                      \STATE {***************\emph{Mutation}****************}
                     \STATE $Offspring$ $\leftarrow$ $Mutation(Parent)$;
                \ENDIF
                \STATE $\mathbb{Q} = \mathbb{Q}\cup Offspring$;
            \ENDFOR
        \ENDFOR
	\end{algorithmic}
\end{algorithm}

Next, we introduce the related prompt designs in the above framework:
\begin{itemize}
  \item \textbf{Prompt for Initialization} (Fig.~\ref{Fig:prompt_initial}): This prompt instructs the LLM to generate initial complete algorithms based on the AMEAD task description. Notably, no prior algorithm information is provided, thereby eliminating the need for expert domain knowledge.
  \item \textbf{Prompt for Inferring Strengths and Drawbacks} (Fig.~\ref{Fig:prompt_drawback}): This prompt supplies the LLM with a component's fitness value, code, and idea, and instructs it to perform an in-depth analysis, summarizing the component's strength and drawback in terms of its practical problem-solving capability.
  \item \textbf{Prompt for Population Selection} (Fig.~\ref{Fig:prompt_population_selection}): This prompt provides the information of all complete algorithms in the archive to the LLM, requiring the LLM to select $N$ algorithms as the initial population for each search. These $N$ algorithms should be of high quality (with proven performance, rigorous logical frameworks, and no fundamental design flaws) and diverse enough to provide rich genetic material for subsequent algorithm evolution.

  \item \textbf{Prompt for Algorithm Recommendation in Lower-Level Search} (Fig.~\ref{Fig:prompt_algorithm_selection}): This prompt elicits a structured chain-of-thought reasoning process to guide the LLM in recommending the most suitable complete algorithm from the archive for pairing with the offspring component during lower-level search: 1) identifying the optimal complete algorithm in the archive as the performance baseline; 2) assessing compatibility between the offspring component and corresponding component of each complete algorithm in the archive; 3) predicting performance gains from component replacement for each algorithm; 4) inferring the optimal target algorithm with maximum expected gain. The fundamental pursuit is to endow the new algorithm after replacement with the potential to outperform the existing optimal algorithm.

\end{itemize}


\subsection{Semantics-Aware Composite Genetic Operator}

To further enable informed and structurally consistent evolution, a semantics-aware composite genetic operator is developed to generate offspring individuals across both the full-algorithm and component search spaces. This operator places semantic understanding at the center of offspring generation, ensuring that crossover and mutation act on meaningful design patterns rather than arbitrary code segments. The pseudocode is given in \textbf{Algorithm~\ref{Alg:Operator}}.

\begin{figure} [!t]
\begin{center}
    \subfigure{\includegraphics[width=0.6\columnwidth]{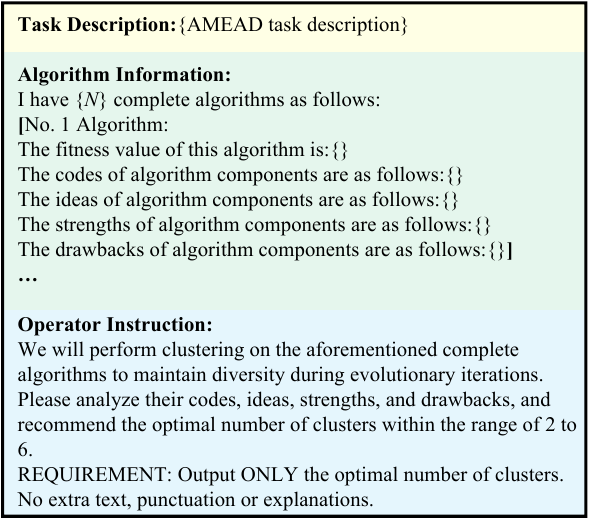}}
    \caption{Prompt for recommending cluster number in upper-level search.}\label{Fig:prompt_global_cluster}
\end{center}
\end{figure}

First, as shown in the prompt in Fig.~\ref{Fig:prompt_global_cluster}, the information of the parent population (denoted as $\mathbb{P}$) is fed into the LLM to prompt it to infer the optimal number of clusters (denoted as $k$). Then, following ParEvo~\cite{ye2024reevo}, we use the code structural semantics as the feature to guide the population clustering. Specifically, for each parent individual, we compute its code similarity with all other parent individuals via CodeBLEU~\cite{ren2020codebleu}, which effectively captures both syntactic structure and semantic consistency of code, yielding a similarity vector. After that, the \emph{K}-means algorithm based on these similarity vectors is performed to partition $\mathbb{P}$ into \emph{k} clusters: $\mathbb{C}_1, \ldots, \mathbb{C}_k$. Consequently, individuals within the same cluster tend to show strong structural consistency and design homogeneity, with similar algorithm architectures and implementation patterns. By contrast, individuals from different clusters are structurally diverse, representing varied and complementary branches of the algorithm design space. This code structural semantics-aware grouping naturally distinguishes local design neighborhoods from global structural variants, providing a reliable basis for targeted and well-balanced genetic operations.

\begin{figure} [!t]
\begin{center}
    \subfigure{\includegraphics[width=0.95\columnwidth]{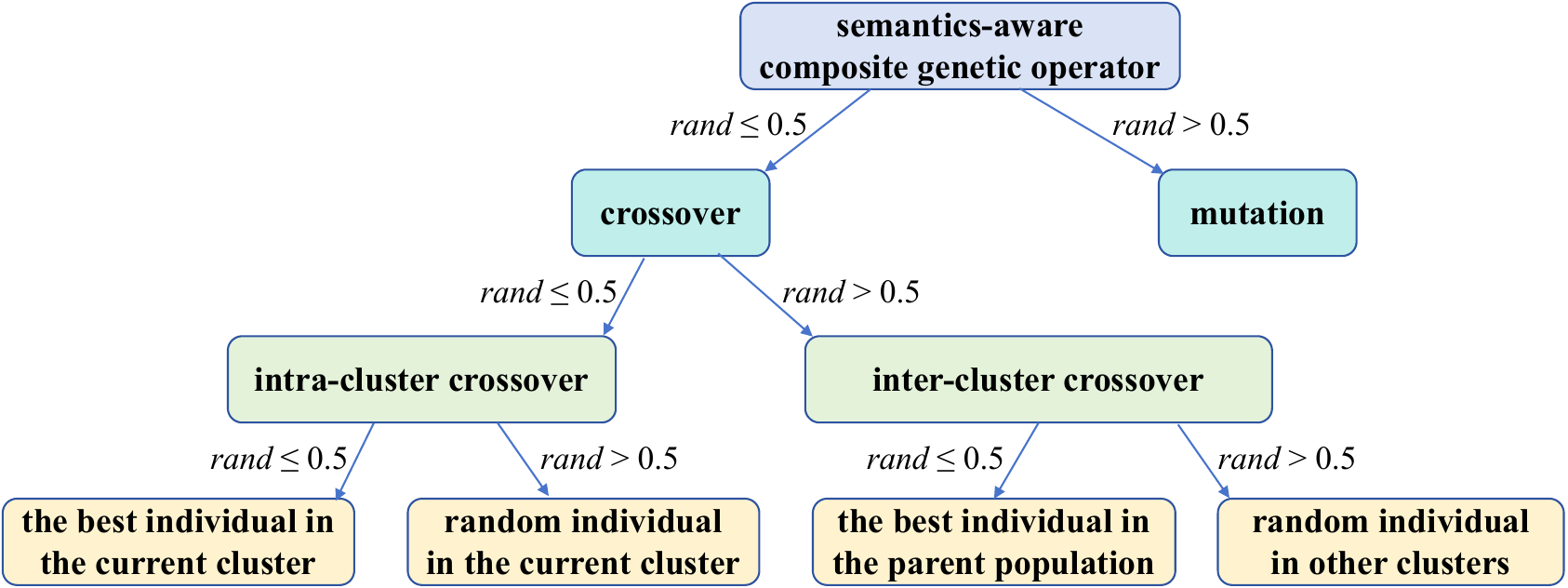}}
    \caption{Semantics-aware composite genetic operator.}\label{Fig:Operator}
\end{center}
\end{figure}

\begin{figure} [!t]
\begin{center}
    \subfigure{\includegraphics[width=0.9\columnwidth]{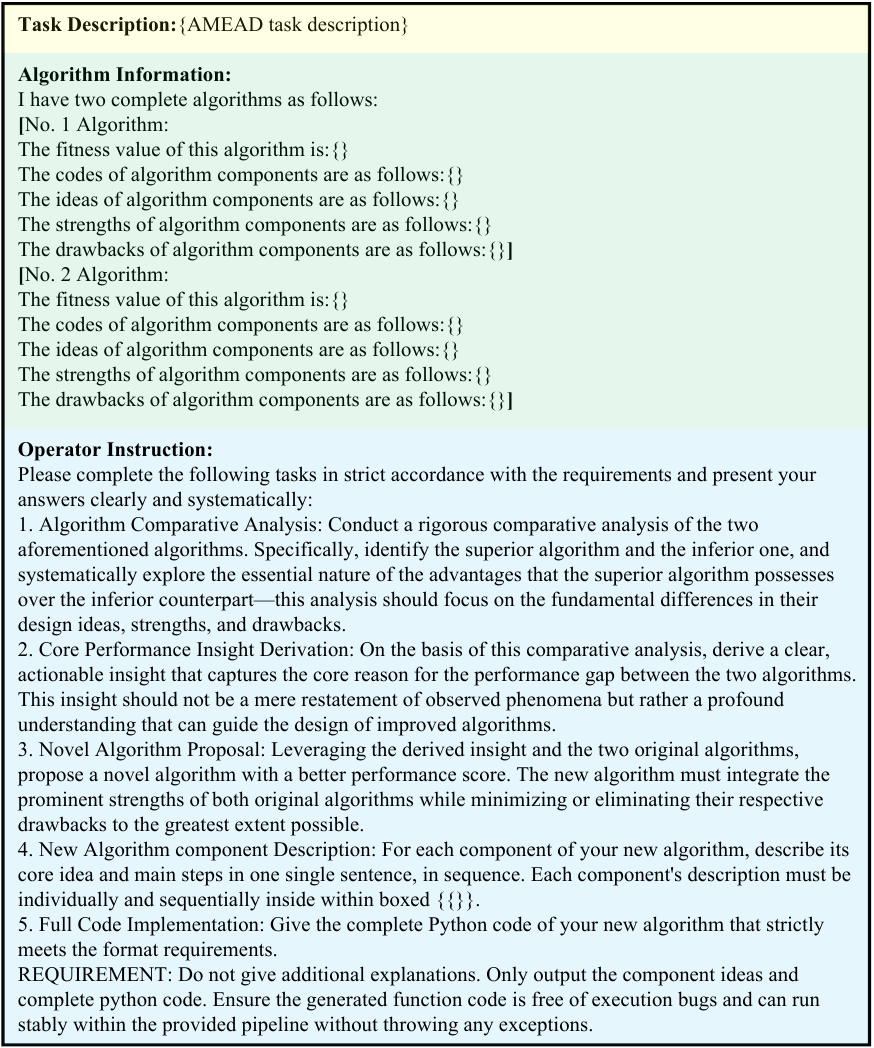}}
    \caption{Prompt for crossover in upper-level search.}\label{Fig:prompt_global_crossover}
\end{center}
\end{figure}

\begin{figure} [!t]
\begin{center}
    \subfigure{\includegraphics[width=0.9\columnwidth]{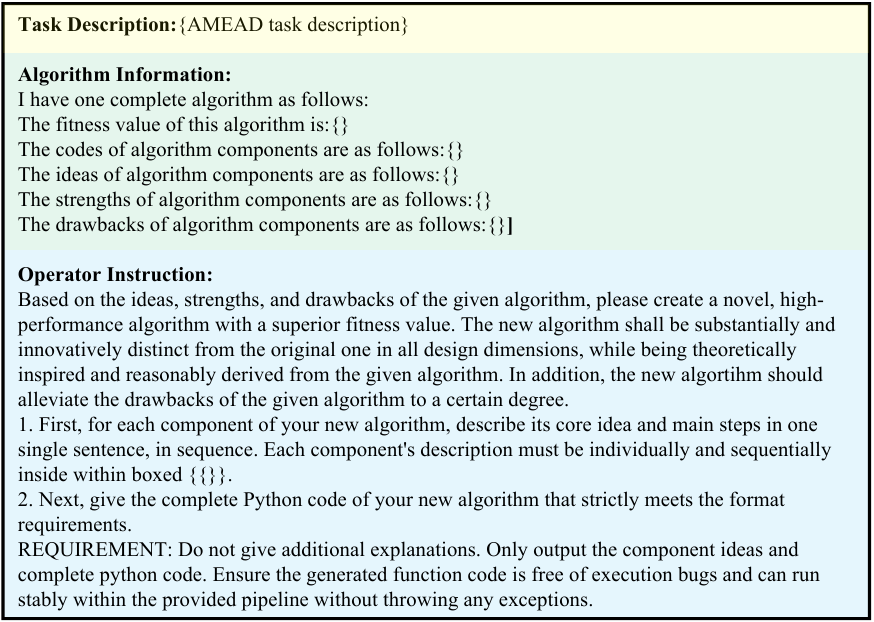}}
    \caption{Prompt for mutation in upper-level search.}\label{Fig:prompt_global_mutation}
\end{center}
\end{figure}

Following population clustering, as shown in Fig.~\ref{Fig:Operator}, each parent individual is subjected to two genetic operators in a probabilistic manner to generate one offspring individual: crossover and mutation. The prompts are given in Fig.~\ref{Fig:prompt_global_crossover} and Fig.~\ref{Fig:prompt_global_mutation}. Crossover semantically integrates design knowledge from two parent individuals to generate functionally improved offspring individuals, while mutation introduces semantically consistent variations that preserve structural rationality rather than disrupting algorithm behavior. Notably, the crossover partner of each parent individual is selected probabilistically from four candidate sources: 1) the best individual in the current cluster, 2) a random individual in the current cluster, 3) the best individual in $\mathbb{P}$, and 4) a random individual from other clusters. The first two choices correspond to intra-cluster crossover, which focuses on semantically refined local exploitation within structurally homogeneous groups. The latter two correspond to inter-cluster crossover, which facilitates semantic knowledge exchange across structurally and functionally distinct regions of the design space. Furthermore, the first and third strategies leverage semantic and performance insights from local and global optima to promote meaningful convergence, whereas the second and fourth inject controlled semantic diversity to avoid premature stagnation. In this way, the proposed semantics-aware composite genetic operator operates at a semantic rather than purely syntactic level, effectively balancing local exploitation and global exploration while maintaining structural and functional coherence throughout the evolutionary process.

Note that in the aforementioned main procedure, an individual represents a complete algorithm in upper-level search, while it denotes an algorithm component in lower-level search. The underlying principles of prompt designs are consistent across both scenarios. Next, we take upper-level search as an example to elaborate on prompt designs.
\begin{itemize}
  \item \textbf{Prompt for Recommending Cluster Number} (Fig.~\ref{Fig:prompt_global_cluster}): This prompt enables the LLM to conduct a multi-dimensional in-depth analysis of the parent population and recommend an appropriate cluster number for the evolutionary process.
  \item \textbf{Prompt for Crossover} (Fig.~\ref{Fig:prompt_global_crossover}): Grounded in the five-dimensional semantics of two parent algorithms, this prompt steers the LLM to conduct semantics-driven chain-of-thought reasoning for principled algorithm generation. It first requires a rigorous comparison to identify performance gaps and distill core semantic advantages from ideas, strengths, and drawbacks. Next, it prompts the derivation of actionable, design-guiding insights into the fundamental causes of performance differences (not superficial descriptions). Finally, using these semantic insights, the LLM synthesizes a high-performance offspring algorithm that integrates parental merits, eliminates critical drawbacks, and maintains structural and functional coherence.
  \item \textbf{Prompt for Mutation} (Fig.~\ref{Fig:prompt_global_mutation}): Driven by the detailed semantic characterizations of the given algorithm, this prompt directs the LLM to perform targeted, semantics-aware innovation. It requires constructing a new algorithm with enhanced fitness, which is not only theoretically grounded in the original algorithm but also exhibits substantial and meaningful innovations across all core design dimensions. Crucially, the generated algorithm must inherently alleviate the critical limitations of the parent algorithm, ensuring all modifications are functionally consistent, semantically valid, and structurally rational.
\end{itemize}
Notably, by explicitly incorporating the strengths and drawbacks of algorithm components, the crossover and mutation prompts establish a clear and knowledge-guided direction for evolutionary improvement: retaining and reinforcing the strengths of the parent individuals while strategically mitigating their drawbacks. This direction  drives continuous and semantics-aware performance enhancement and ensuring the stability, interpretability, and effectiveness of the entire evolutionary process.

\emph{Remark} 1: We also show the prompts for recommending cluster number, crossover, and mutation in lower-level search in Figs. S-1--S-3 of the supplementary file.

\subsection{Semantics-Aware Performance Evaluator}
\begin{algorithm}[!t]
	\renewcommand{\algorithmicrequire}{\textbf{Input:}}
	\renewcommand{\algorithmicensure}{\textbf{Output:}}
	\caption{Semantics-Aware Performance Evaluator}\label{Alg:Evaluator}
	\begin{algorithmic}[1]\footnotesize
		\REQUIRE parent individuals, offspring individual, $APE_\textup{con}$, and $APE_{\max}$
        \ENSURE \emph{fitness value} of the offspring individual
        \STATE  Calculate the control probability $Control\_Pro$ according to (\ref{Eqn:Pro});
        \IF{$rand > Control\_Pro$}
		\STATE  \emph{fitness value} of the offspring individual $\leftarrow$ performance evaluation on the pre-defined optimization instances;
        \ELSE
        \STATE  \emph{fitness value} of the offspring individual $\leftarrow$ fitness inheritance based on \emph{fitness values} of the parent individuals;
        \ENDIF
	\end{algorithmic}
\end{algorithm}

To enable the rational allocation of limited computational resources across different regions of the design space and to facilitate deeper and broader search, we introduce a semantics-aware performance evaluator, which consists of two evaluation modes: 1) true evaluation and 2) fitness inheritance. In the evolutionary computation community, fitness inheritance exploits historical information to estimate the fitness values of new individuals without true evaluations, acting as an efficiency-enhancing technique to accelerate optimization under limited computational resources~\cite{chen2002fitness}. The selection between the two modes is determined by the code structural semantics of the offspring individuals, as shown in \textbf{Algorithm~\ref{Alg:Evaluator}}.

To be specific, when evaluating an offspring individual, we first calculate a control probability:
		\begin{equation}\label{Eqn:Pro}
		\begin{aligned}
			Control\_Pro = Code\_Sim \cdot \frac{1}{1+\exp(-\frac{1-APE_\textup{con}/APE_{\max}}{2})}
		\end{aligned}
		\end{equation}
In (\ref{Eqn:Pro}), $APE_\textup{con}$ denotes the current consumed number of APEs, and $APE_{\max}$ denotes the maximum number of available APEs. In addition, in upper-level search, $Code\_Sim$ represents the mean code structural similarity between the offspring algorithm and two crossover parent algorithms, or the code structural similarity between the offspring algorithm and the mutation parent algorithm. In lower-level search, $Code\_Sim$ is defined similarly, but computed between the original complete algorithms of parent components and the combination of the offspring component with its LLM-recommended complete algorithm, covering both crossover and mutation scenarios. Here, the code structural similarity between any two complete algorithms is calculated based on CodeBLEU.

Then, a random number is sampled from the interval [0, 1]. If this random number exceeds $Control\_Pro$, the first evaluation mode is triggered: the performance of the offspring individual is evaluated on the pre-defined optimization instances, with one APE consumed. Conversely, if the random number is smaller than $Control\_Pro$, the second evaluation mode is activated, where the fitness value of the offspring individual is assigned to that of a randomly selected crossover parent or the mutation parent, without incurring one APE.

Based on the above introduction, we can draw two observations about (\ref{Eqn:Pro}), i.e., $Control\_Pro$:
\begin{enumerate}
  \item $Control\_Pro$ is positively correlated with $Code\_Sim$. Higher structural similarity between parents and offspring leads to a higher probability of fitness inheritance. This mechanism enables reliable fitness inheritance for structurally similar individuals, reduces valuable computational resources consumed on structure-semantics similar areas in the design space, and instead encourages broader exploration towards unexploited areas enables trustworthy fitness inheritance.

  \item  $Control\_Pro$ decreases monotonically with increasing $APE_\textup{con}$. In the early stage, higher $Control\_Pro$ promotes fitness inheritance to reduce APE consumption and support broad global exploration; in the later stage, lower $Control\_Pro$ encourages true evaluation for precise assessment local exploitation, effectively prolonging evolutionary generations to strengthen search depth.
\end{enumerate}
Combining the above two observations, the proposed semantics-aware performance evaluator effectively adaptively allocates limited computational resources across different regions of the design space in a reasonable manner, reduces redundant APE consumption throughout evolution, and thereby facilitates deeper and broader search, ultimately accelerating the overall evolutionary process.

\section{Experimental Studies}\label{Sec:Set}
In this section, we verify the effectiveness of STABLE through two sets of experiments. We first introduce the two defined AMAD tasks in Section \ref{Sec:E1}, followed by the experimental settings of two sets of experiments in Section \ref{Sec:E2}. Then, Sections \ref{Sec:E3} and \ref{Sec:E4} give the discussion of experimental results.

\subsection{Two AMAD Tasks}\label{Sec:E1}
We implement two AMAD tasks: constrained MOEAs~\cite{10897802,10518071,11475039} and surrogate-assisted MOEAs~\cite{10379505,11283053,10509608}. The task descriptions are shown in Figs.~S-4 and S-5 of the supplementary file. The component design task descriptions can be easily extracted from the overall task descriptions. In addition, the basic introductions of constrained and surrogate-assisted MOEAs are given in Section S-III of the supplementary file.

To ensure a fair comparison, the same initial population is adopted when using different LES methods to tackle the two AMAD tasks. Specifically, for the constrained MOEA design task, we consider 15 constrained MOEAs constructed by combining three mainstream MOEAs (i.e., NSGA-II~\cite{996017}, MOEA/D~\cite{zhang2007moea}, and IBEA~\cite{zitzler2004indicator}) and five classic constraint-handling techniques (i.e., constrained-domination principle, multiobjective-based method, $\varepsilon$-constrained method, self-adaptive penalty-based method, and stochastic ranking~\cite{9440869}). For the surrogate-assisted MOEA design task, the initial 15 surrogate-assisted MOEAs are formed by integrating NSGA-II, MOEA/D, and IBEA with five widely-used surrogate models, including the Kriging model, radial basis function, polynomial response surface, artificial neural network, and support vector regression~\cite{jin2005comprehensive}. The above initial algorithms can provide abundant and diverse strategy genes for evolution.


In the constrained MOEA design task, 10-decision-variable MW4, MW5, MW9, and MW10~\cite{8632683} are used to evaluate the performance of constrained MOEAs. MW4 has 3 objectives, while the other three MW test problems are 2-objective. For each constrained MOEA, the maximum number of available fitness evaluations for each test problem is set to 10000, and its final fitness value is defined as the negative average Hypervolume~\cite{zitzler2004indicator} (NAHV) of feasible nondominated solutions over the four test problems. Similarly, in the surrogate-assisted MOEA design task, 3-objective, 10-decision-variable DTLZ1, DTLZ2, DTLZ4, and DTLZ7~\cite{deb2005scalable} with relaxed $g$ functions are used. For each surrogate-assisted MOEA, the maximum number of available fitness evaluations for each test problem is set to 60, and its final fitness value is also defined as the NAHV. Note that the smaller the NAHV, the better the convergence and diversity of the obtained solutions.
\begin{table}[!t]
  \centering
  \caption{Mean NAHV values Over Three Runs Provided by EoH, ParEvo, STABLE, and Four STABLE Variants on the Two AMAD Tasks.}
    \begin{tabular}{ccc}
    \hline
    \multirow{2}[2]{*}{LES methods} & \multicolumn{1}{c}{Constrained MOEA} & Surrogate-Assisted MOEA  \bigstrut[t]\\
          & Design Task & Design Task \bigstrut[b]\\
    \hline
    EoH   & -0.553  & -0.557  \bigstrut[t]\\
    ParEvo & -0.573  & -0.571  \\
    STABLE\_global & -0.648  & -0.592  \\
    STABLE\_local & -0.603  & -0.591  \\
    STABLE\_repre & -0.584  & -0.655  \\
    STABLE\_fitness & -0.600  & -0.648  \\
    STABLE & \textbf{-0.673} & \textbf{-0.687} \bigstrut[b]\\
    \hline
    \end{tabular}%
  \label{tab:result1}%
\end{table}%

    \begin{figure*} [!t]
		\begin{center}
        \subfigure[]{\label{Fig:Convergence1}\includegraphics[width=0.9\columnwidth]{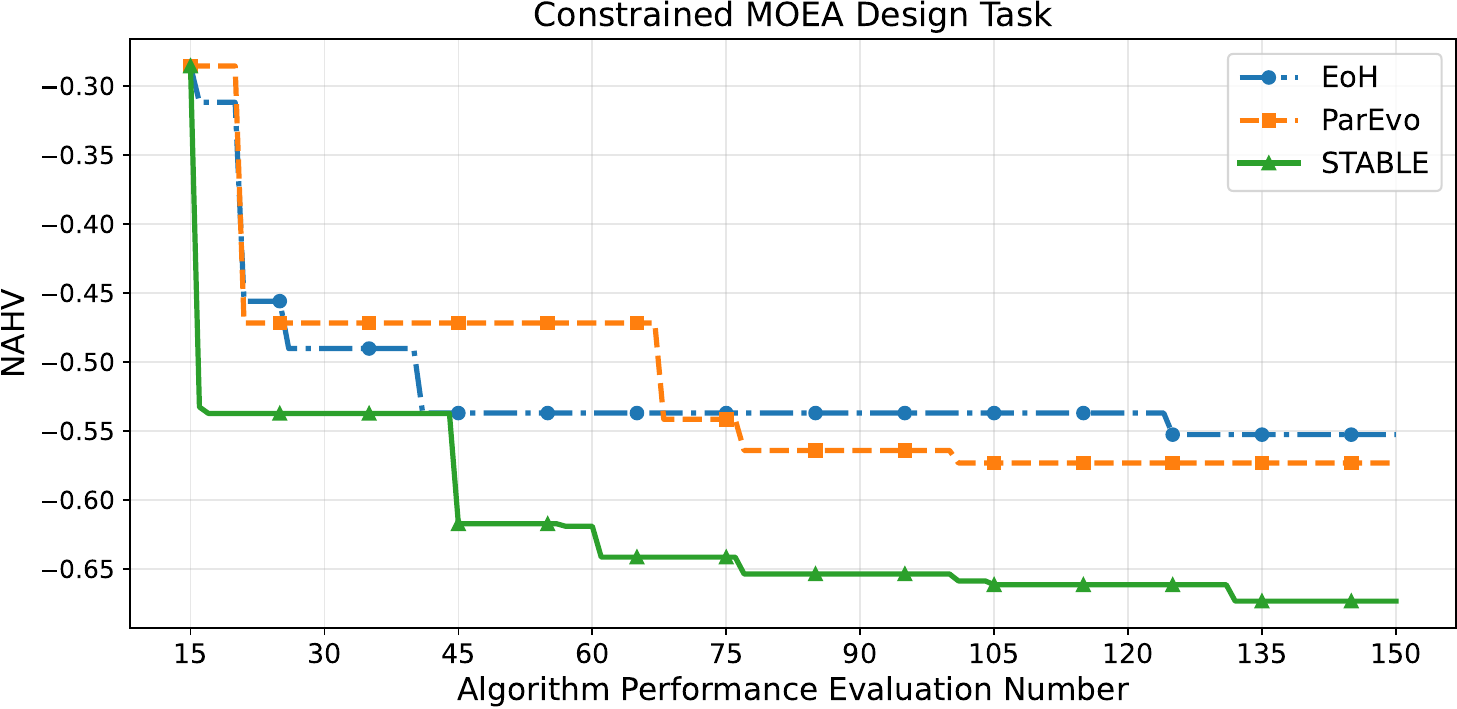}}
        \subfigure[]{\label{Fig:Convergence2}\includegraphics[width=0.9\columnwidth]{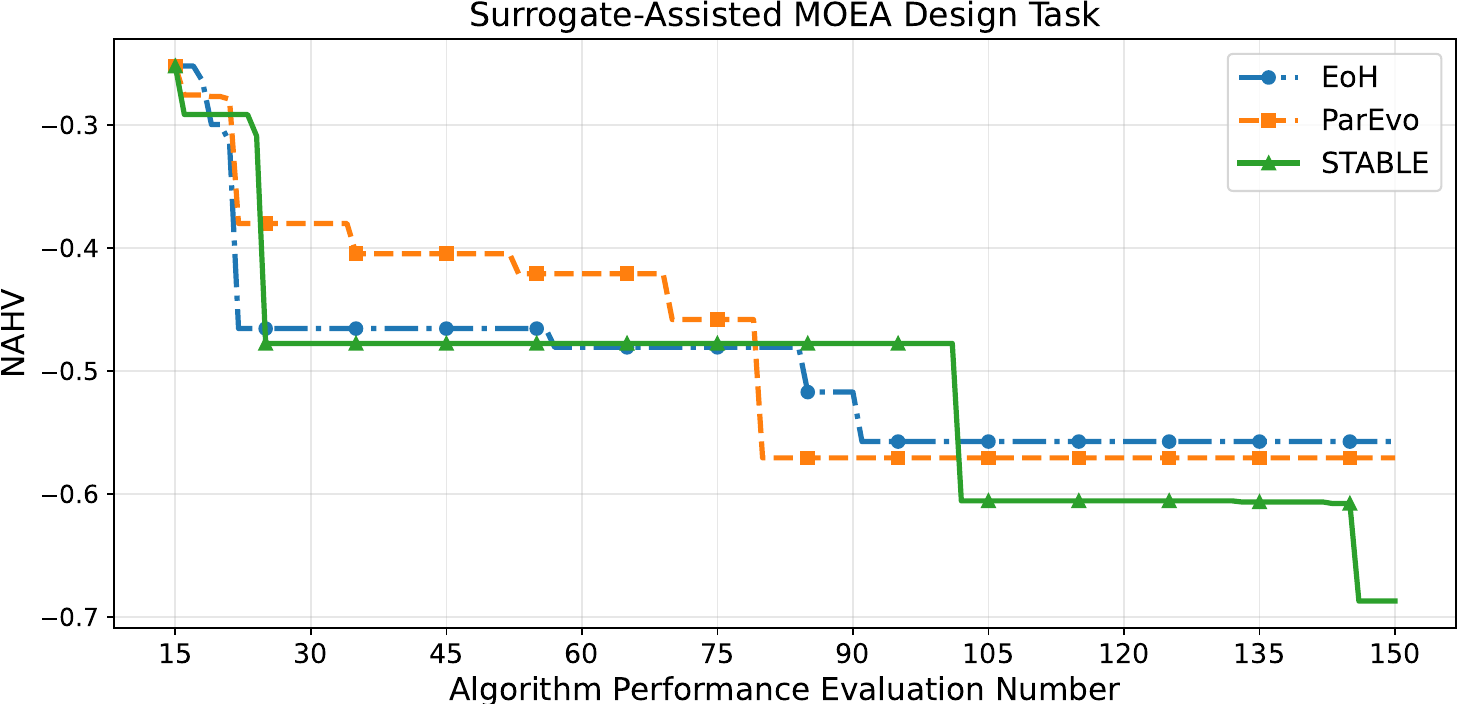}}
			\caption{Convergence curve of the median NAHV values over three runs provided by EoH, ParEvo, and STABLE on (a) the constrained MOEA design task and (b) the surrogate-assisted MOEA design task.}\label{Fig:Conver1}
		\end{center}
	\end{figure*}

    \begin{figure*} [!t]
		\begin{center}
        \subfigure[]{\label{Fig:Convergence3}\includegraphics[width=0.9\columnwidth]{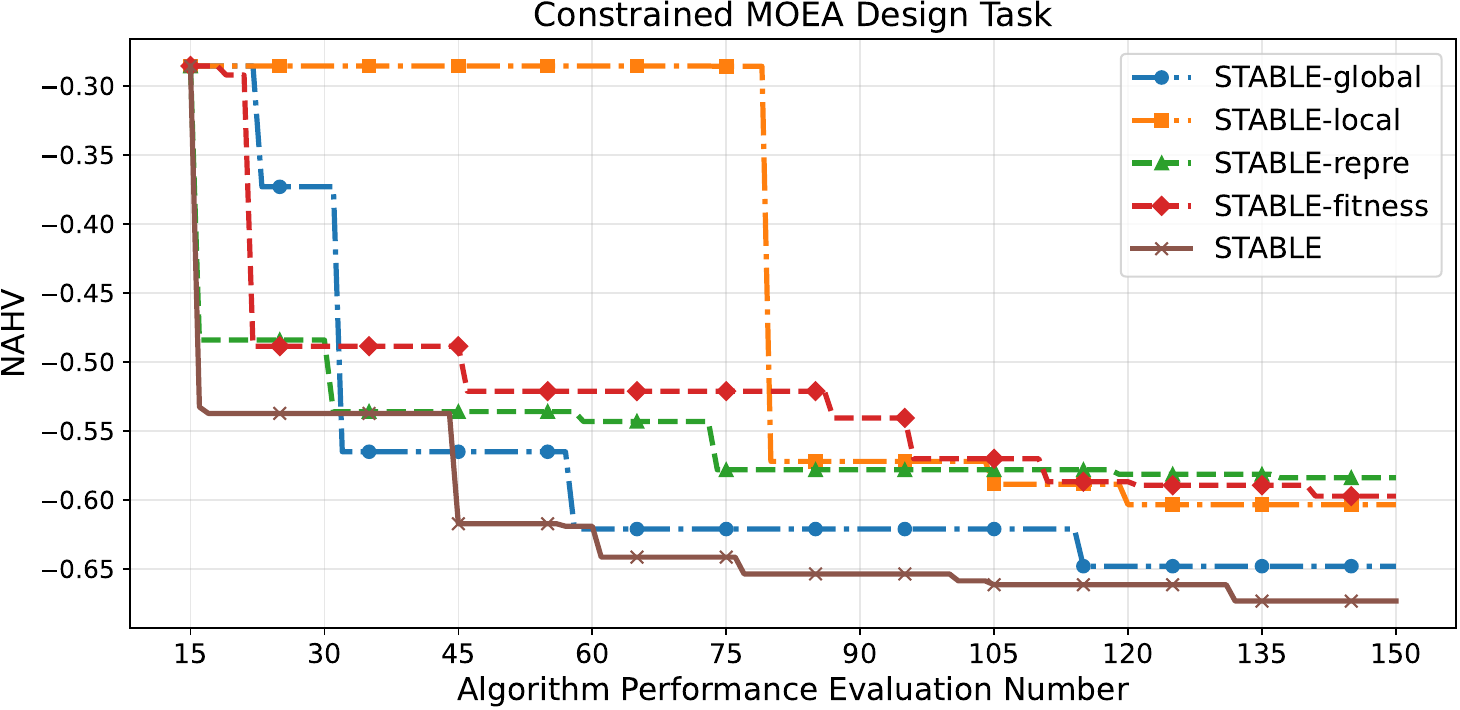}}
        \subfigure[]{\label{Fig:Convergence4}\includegraphics[width=0.9\columnwidth]{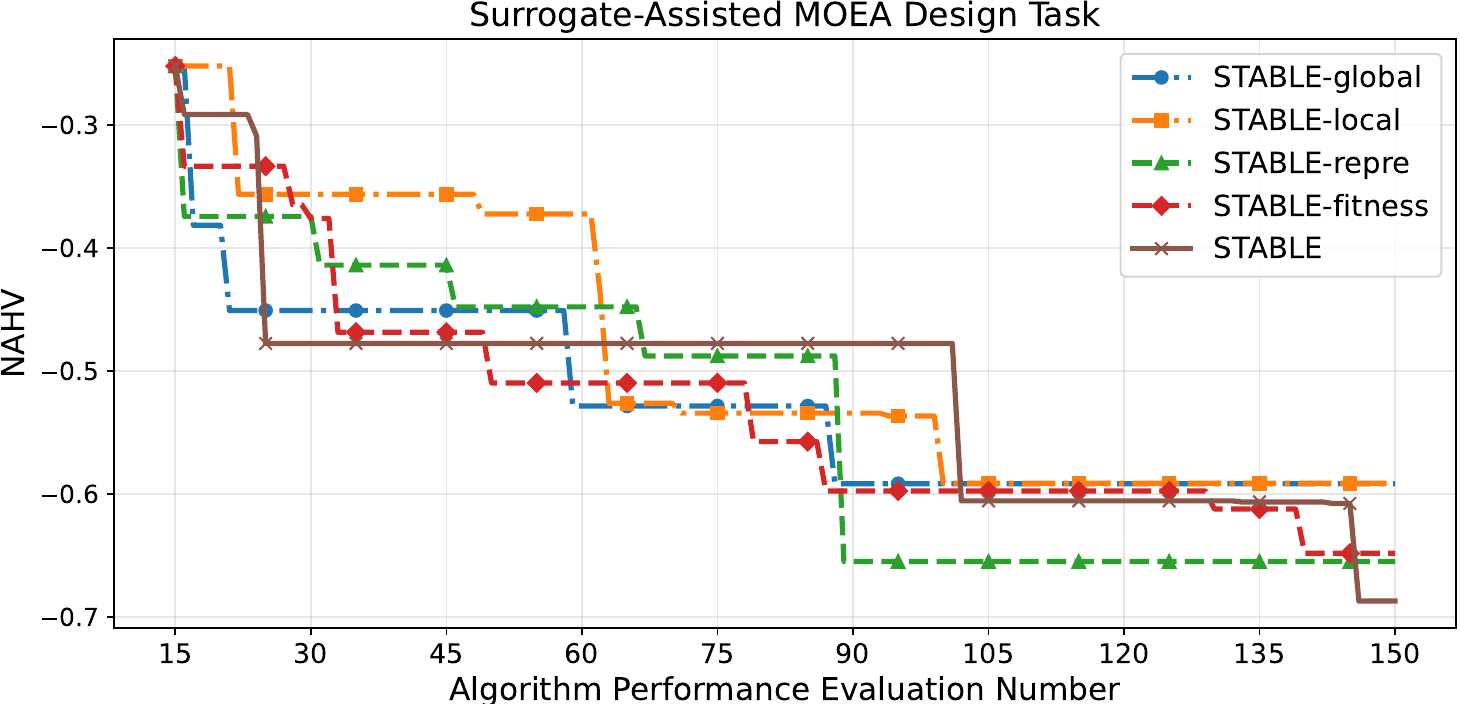}}
			\caption{Convergence curve of the median NAHV values over three runs provided by STABLE and its four variants on (a) the constrained MOEA design task and (b) the surrogate-assisted MOEA design task.}\label{Fig:Conver}
		\end{center}
	\end{figure*}	

\subsection{Experimental Settings of Two Sets of Experiments}\label{Sec:E2}
We verify the effectiveness of STABLE through  two sets of experiments: 1) performance evaluation of STABLE on two AMAD tasks; 2) performance investigation of the best algorithms designed by STABLE.

In the first set of experiments, we compare STABLE with two representative LES methods: EoH~\cite{liu2024evolution} and ParEvo~\cite{hu2025partition}. They have already been introduced in Section~\ref{Sec:Rew}. To alleviate the inherent randomness in LLM responses, the three LES methods are executed three times independently for each AMAD task. In addition, for all three LES methods, the population size is set to $N$ = 15, the maximum number of available APEs is $APE_{\max}$ = 150, and the employed commercial LLM is DeepSeek-V3.2~\cite{liu2025deepseek}. All algorithms generated by the three LES methods are implemented in Python. In addition, we also conduct ablation studies to investigate the effectiveness of main components in STABLE.

In the second set of experiments, on the one hand, we compare the best constrained MOEAs designed by STABLE, EoH, and ParEvo (denoted as CMOEA-STABLE, CMOEA-EoH, and CMOEA-ParEvo, respectively) with CCMO~\cite{9122020}, $\theta$-DEA-CPBI~\cite{ming2023constraint}, Ship~\cite{9440869}, DRLOS-EMCMO~\cite{ming2024constrained}, and DVCEA~\cite{ban2025decision} on MW~\cite{8632683} and CTP~\cite{deb2001constrained} test suites. For each constrained MOEA, the population size is set to 100, the maximum number of allowed fitness evaluations is set to 30000, and the number of independent runs is set to 20. On the other hand, the best surrogate-assisted MOEAs designed by STABLE, EoH, and ParEvo (denoted as SAMOEA-STABLE, SAMOEA-EoH, and SAMOEA-ParEvo, respectively) are compared with K-RVEA~\cite{7723883}, EDN-ARMOEA~\cite{9312457}, MCEA/D~\cite{9733428}, SSDE~\cite{araujo2024self}, and DISK~\cite{10804681} on DTLZ~\cite{deb2005scalable} and UF~\cite{zhang2008multiobjective} test suites. For each surrogate-assisted MOEA, the initial database size and population size are both set to 50, the maximum number of available fitness evaluations is limited to 200, and 20 independent runs are conducted. In the above four test suites, the number of objectives (i.e., $m$) for MW4, MW8, MW14, all DTLZ test problems, and UF8--UF10 is 3, and the remaining test problems are 2-objective, and the number of decision variables for all test problems is set to $D=10$. In addition, the modified inverted generational distance (IGD$^+$)~\cite{ishibuchi2015modified} is used to measure the quality of (feasible) nondominated solutions obtained by the above algorithms, with a smaller value indicating better convergence and diversity. Note that the above comparison experiments were conducted on the PlatEMO~\cite{tian2017platemo} platform. To ensure consistency, the algorithms produced by the three LES methods were converted from Python to MATLAB before execution.

\emph{Remark}~2: The human-designed competitors and test suites in the second set of experiments are introduced in Section S-IV of the supplementary file.

\subsection{Performance Evaluation of STABLE on Two AMAD Tasks}\label{Sec:E3}

\subsubsection{Comparison with Advanced LES Methods}
The results of EoH, ParEvo, and STABLE on the two AMAD tasks are given in Table \ref{tab:result1}. From Table \ref{tab:result1}, the mean NAHV values over three runs provided by EoH, ParEvo, and STABLE are -0.553, -0.573, and -0.673 on the constrained MOEA design task, respectively, and are -0.557, -0.571, and -0.687 on the surrogate-assisted MOEA design task, respectively. Clearly, the results reported by STABLE are far better than those of EoH and ParEvo, showing its stronger capability in handling AMAD tasks.

Fig.~\ref{Fig:Conver1} depicts the convergence curves of the median NAHV values over three runs provided by EoH, ParEvo, and STABLE on the two AMAD tasks. The horizontal axis represents the algorithm performance evaluation number, which is capped at a maximum value of 150. It can be observed that the NAHV values start at a relatively high level, approximately between -0.25 and -0.30. As evolution proceeds, the quality of the designed algorithms is gradually refined, demonstrating the effectiveness of the LES methods. As shown in Fig.~\ref{Fig:Convergence1}, STABLE consistently outperforms EoH and ParEvo in terms of convergence speed from the outset on the constrained MOEA design task. Notably, EoH and ParEvo suffer from severe convergence stagnation during the early and middle stages, which likely stems from diminished search efficiency in the large search spaces of AMAD tasks. From Fig.~\ref{Fig:Convergence2}, on the surrogate-assisted MOEA design task, EoH, ParEvo, and STABLE exhibit similarly slow search trajectories in the early and middle stages. This can be attributed to the fact that the five-component search space imposes significantly greater search difficulty for the three LES methods under a limited APE budget. However, STABLE evolves considerably faster in the later stages, owing to its lower-level strategy, which effectively refines the quality of the current potential algorithms.

Overall, the above results demonstrate that STABLE constitutes a more powerful LES approach than EoH and ParEvo for AMAD. This can be attributed to two factors: 1) the bilevel co-evolutionary framework effectively decouples the complex hierarchical design spaces of AMAD tasks, attains a well-balanced tradeoff between exploration and exploitation, and consequently yields superior optimization performance; 2) the five-dimensional semantic model provides abundant prior knowledge, allowing for targeted and effective guidance throughout algorithm generation and evaluation.

\subsubsection{Ablation Studies}
Table \ref{tab:result1} also records the results of STABLE and its four variants: STABLE-global, STABLE-local, STABLE-repre, and STABLE-fitness. STABLE-global only implements upper-level search and STABLE-local only executes lower-level search. In STABLE-repre, each algorithm component is only characterize by three-dimensional semantics: code, idea, and fitness value. In STABLE\_fitness, the semantics-aware performance evaluator only implements true evaluation. As shown in Table \ref{tab:result1}, the mean NAHV values of STABLE are smaller than the four variants on the two AMAD tasks.

Fig.~\ref{Fig:Conver} illustrates the convergence curves of median NAHV values across three independent runs for STABLE-global, STABLE-local, STABLE-repre, STABLE-fitness, and STABLE on the two AMAD tasks. As can be observed in Fig.~\ref{Fig:Conver}, STABLE-global displays convergence behavior similar to STABLE in the early stages, yet achieves a slower convergence rate in later phases. This arises because lower-level search in STABLE facilitates stronger exploitation around the current promising algorithms. In contrast, STABLE-local exhibits noticeable convergence stagnation or slow progress in the early stages. The underlying reason is that the initial algorithms in the archive are mostly low-performance, and lower-level search primarily refines these inferior solutions, resulting in slow initial convergence. Furthermore, both STABLE-repre and STABLE-fitness also yield lower convergence speeds compared to STABLE.

In summary, the above results fully validate the effectiveness of the bilevel co-evolutionary framework in balancing exploration and exploitation, as well as the benefits of the five-dimensional semantic model for significantly enhancing search efficiency.

\begin{figure*} [!t]
\begin{center}
    \subfigure{\includegraphics[width=1.6\columnwidth]{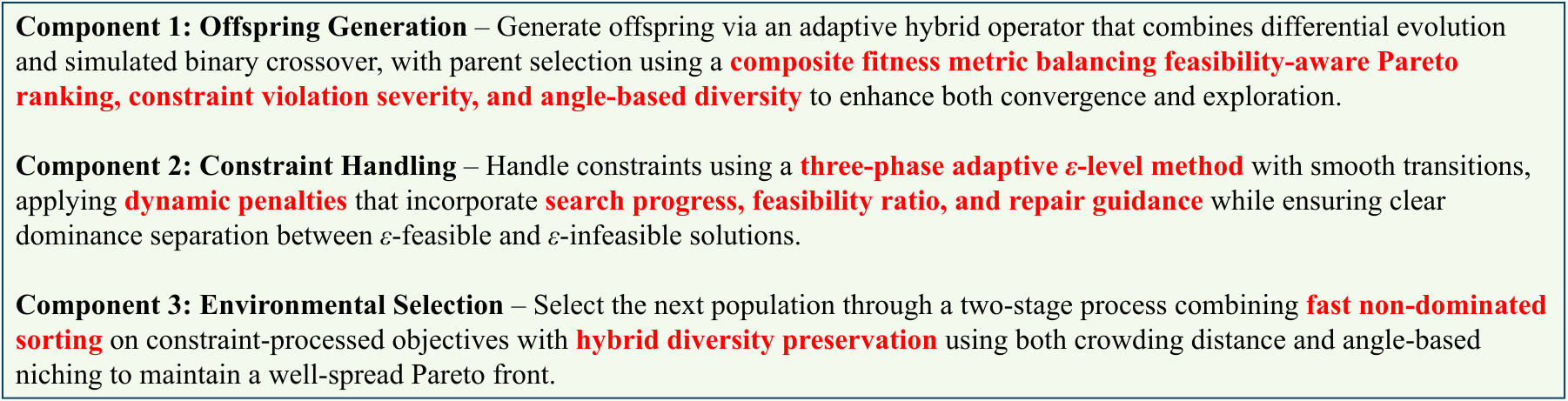}}
    \caption{Ideas of the three components of the best constrained MOEA designed by STABLE, i.e., CMOEA-STABLE.}\label{Fig:CMOEA-STABLE}
\end{center}
\end{figure*}

\begin{figure*} [!t]
\begin{center}
    \subfigure{\includegraphics[width=1.6\columnwidth]{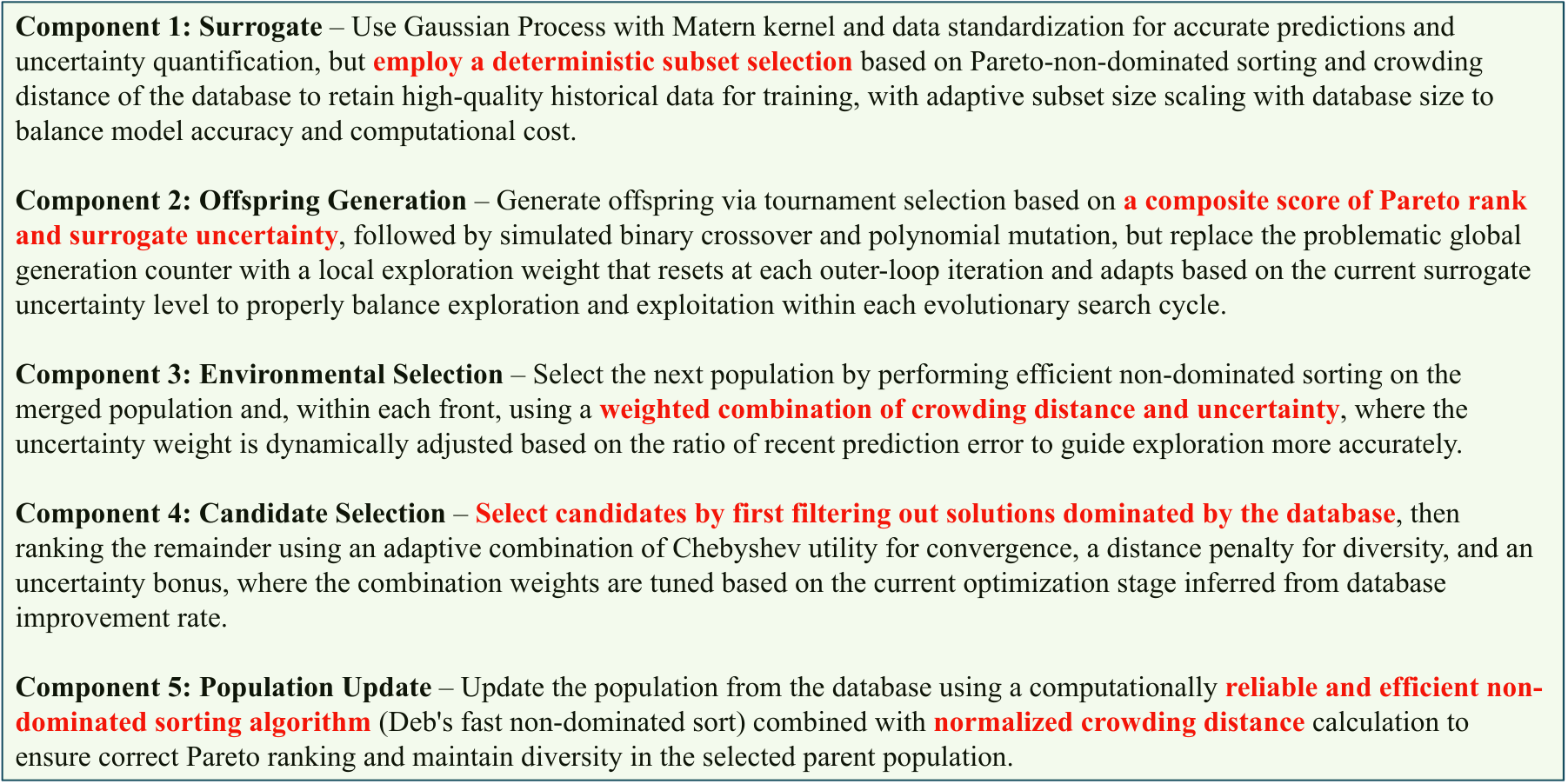}}
    \caption{Ideas of the five components of the best surrogate-assisted MOEA designed by STABLE, i.e., SAMOEA-STABLE.}\label{Fig:SAMOEA-STABLE}
\end{center}
\end{figure*}

\subsubsection{Best Algorithms Designed by STABLE}
To intuitively demonstrate the algorithm design capability of STABLE for AMAD tasks, we briefly show the best constrained and surrogate-assisted MOEAs discovered by STABLE, denoted as CMOEA-STABLE and SAMOEA-STABLE, respectively. Due to the excessive length of the codes, we only present the ideas, as shown in Figs. \ref{Fig:CMOEA-STABLE} and \ref{Fig:SAMOEA-STABLE}.

From Figs. \ref{Fig:CMOEA-STABLE} and \ref{Fig:SAMOEA-STABLE}, the best algorithms designed by STABLE adaptively integrate advanced experiences and state-of-the-art mechanisms from constrained and surrogate-assisted MOEA research, flexibly assembling mature strategies including adaptive constraint handling via three-phase $\varepsilon$-level methods with dynamic penalty schemes, diversity preservation through hybrid crowding-distance and angle-based niching, nondominated sorting with constraint-aware fast ranking, and surrogate-guided search leveraging Gaussian process models with uncertainty-aware selection in a task-oriented way. By reasonably inheriting, reorganizing, and synergizing the core design paradigms accumulated in the existing literature, the generated algorithm architectures exhibit strong rationality, effectiveness, and task adaptability. This fully reflects the rich domain-specific knowledge embedded in LLMs during pre-training, which enables them to effectively master and utilize mainstream evolutionary computation methodologies, providing solid knowledge support for AMAD tasks. Furthermore, STABLE serves as an effective framework to excavate and activate such inherent prior knowledge within LLMs, guiding the model to transfer and apply domain insights into practical algorithm design and further unleash the potential of LLMs in AMAD.

\emph{Remark}~3: The ideas of the best algorithms designed by EoH and ParEvo (i.e., CMOEA-EoH, SAMOEA-EoH, CMOEA-ParEvo, and SAMOEA-ParEvo) are shown in Figs. S-8--S-11 of the supplementary file.

\subsection{Performance Investigation of the Best Algorithms Designed by STABLE}\label{Sec:E4}
Tables \ref{tab:RE1}--\ref{tab:RE4} present the experimental results of CMOEA-STABLE/SAMOEA-STABLE and its competitors on the test suites in terms of $\textup{IGD}^{+}$, where the averages and standard deviations are recorded. To evaluate the statistical significance regarding the $\textup{IGD}^{+}$ metric, the Wilcoxon rank-sum test was performed at a 0.05 significance level between CMOEA-STABLE/SAMOEA-STABLE and each compared algorithm. ``$+$''/``$-$'' denotes that the competitor outperforms or underperforms CMOEA-STABLE/SAMOEA-STABLE, respectively, and ``$\approx$'' indicates that the competitor achieves statistically similar performance to CMOEA-STABLE/SAMOEA-STABLE. For each test problem, the best result is highlighted in boldface and green, and the second-best result is highlighted in yellow.

\begin{table*}[!t]
  \centering
  \caption{Results on MW test Suite.}
    \resizebox{18cm}{!}{\begin{tabular}{|c|c|c|c|c|c|c|c|c|c|c|}
    \hline
    Problem & \textit{m} & \textit{D} & CCMO  & $\theta$-DEA-CPBI & ShiP  & DRLOS-EMCMO & DVCEA & CMOEA-EoH & CMOEA-ParEvo & CMOEA-STABLE \bigstrut\\
    \hline
    MW1   & \multirow{3}[6]{*}{2} & \multirow{14}[28]{*}{10} & 2.4972e-3 (4.35e-5) $-$ & 9.8815e-3 (1.76e-2) $-$ & 4.1686e-3 (6.47e-3) $-$ & 2.8036e-3 (1.11e-4) $-$ & \cellcolor{yellow!30}2.3578e-3 (3.12e-5) $-$ & 1.0608e-1 (8.67e-2) $-$ & 1.2688e-1 (2.33e-1) $\approx$ & \cellcolor{green!30}\textbf{1.3124e-3 (3.28e-5)} \bigstrut\\
\cline{1-1}\cline{4-11}    MW2   &       &       & 1.9248e-2 (8.87e-3) $\approx$ & 2.2178e-2 (1.45e-2) $-$ & \cellcolor{green!30}\textbf{1.3748e-2 (1.06e-2) $\approx$} & 2.6827e-2 (9.44e-3) $-$ & 3.0374e-2 (1.10e-2) $-$ & 1.2141e-1 (1.14e-1) $-$ & 8.2114e-2 (6.17e-2) $-$ & \cellcolor{yellow!30}1.4228e-2 (7.72e-3) \bigstrut\\
\cline{1-1}\cline{4-11}    MW3   &       &       & 5.6787e-3 (5.70e-4) $-$ & 4.5211e-2 (1.66e-1) $-$ & 7.1742e-3 (1.45e-2) $-$ & 6.3150e-3 (4.26e-4) $-$ & \cellcolor{yellow!30}4.6297e-3 (2.70e-4) $-$ & 8.8347e-2 (9.40e-2) $-$ & 5.0365e-3 (5.12e-4) $-$ & \cellcolor{green!30}\textbf{3.4176e-3 (3.06e-4)} \bigstrut\\
\cline{1-2}\cline{4-11}    MW4   & 3     &       & 4.3405e-2 (6.41e-4) $+$ & 4.3833e-2 (1.07e-4) $\approx$ & \cellcolor{yellow!30}4.0332e-2 (2.23e-3) $+$ & 4.6349e-2 (1.53e-3) $-$ & 4.3231e-2 (6.65e-4) $+$ & 2.6064e-1 (1.28e-1) $-$ & \cellcolor{green!30}\textbf{3.7379e-2 (4.44e-3) $+$} & 4.4621e-2 (1.95e-3) \bigstrut\\
\cline{1-2}\cline{4-11}    MW5   & \multirow{3}[6]{*}{2} &       & \cellcolor{yellow!30}8.1600e-3 (1.25e-2) $+$ & 7.0216e-2 (1.04e-1) $$ & 4.1314e-2 (7.34e-2) $-$ & \cellcolor{green!30}\textbf{2.5846e-3 (9.88e-4) $+$} & 9.0773e-3 (2.34e-3) $+$ & 2.1409e-1 (9.55e-2) $-$ & 1.7325e-1 (1.50e-1) $-$ & 3.7517e-2 (2.11e-2) \bigstrut\\
\cline{1-1}\cline{4-11}    MW6   &       &       & \cellcolor{yellow!30}2.3708e-2 (2.23e-2) $\approx$ & 4.4616e-2 (8.16e-2) $-$ & 5.3869e-2 (1.09e-1) $\approx$ & 3.5761e-2 (2.27e-2) $-$ & 5.0297e-2 (4.19e-2) $-$ & 3.3934e-1 (1.82e-1) $-$ & 4.2305e-1 (2.53e-1) $-$ & \cellcolor{green!30}\textbf{1.8398e-2 (1.26e-2)} \bigstrut\\
\cline{1-1}\cline{4-11}    MW7   &       &       & 3.7998e-3 (3.52e-4) $-$ & 2.0174e-2 (6.71e-2) $-$ & 1.8718e-2 (6.77e-2) $\approx$ & 4.2557e-3 (2.36e-4) $-$ & \cellcolor{yellow!30}3.6096e-3 (1.38e-4) $-$ & 1.2887e-1 (5.47e-2) $-$ & 8.0011e-3 (9.17e-4) $-$ & \cellcolor{green!30}\textbf{3.1857e-3 (2.17e-4)} \bigstrut\\
\cline{1-2}\cline{4-11}    MW8   & 3     &       & 4.2880e-2 (1.16e-2) $\approx$ & 5.6355e-2 (3.82e-2) $-$ & \cellcolor{green!30}\textbf{3.2480e-2 (5.54e-3) $+$} & 5.0462e-2 (1.10e-2) $-$ & 5.2042e-2 (1.09e-2) $-$ & 2.9257e-1 (1.16e-1) $-$ & 8.1958e-2 (2.55e-2) $-$ & \cellcolor{yellow!30}3.8172e-2 (7.50e-3) \bigstrut\\
\cline{1-2}\cline{4-11}    MW9   & \multirow{5}[10]{*}{2} &       & 7.1802e-2 (2.00e-1) $-$ & 1.8362e-2 (1.16e-2) $\approx$ & 3.0130e-2 (1.21e-2) $-$ & \cellcolor{green!30}\textbf{7.6781e-3 (1.12e-3) $+$} & \cellcolor{yellow!30}9.5042e-3 (1.88e-3) $+$ & 6.9240e-1 (5.08e-2) $-$ & 5.0296e-2 (1.30e-1) $\approx$ & 2.0520e-2 (8.49e-3) \bigstrut\\
\cline{1-1}\cline{4-11}    MW10  &       &       & 5.9904e-2 (5.11e-2) $\approx$ & 9.7174e-2 (8.40e-2) $-$ & \cellcolor{yellow!30}4.8739e-2 (3.64e-2) $\approx$ & 7.0539e-2 (6.21e-2) $-$ & 9.0288e-2 (6.95e-2) $-$ & 2.2218e-1 (8.71e-2) $-$ & 2.6480e-1 (9.18e-2) $-$ & \cellcolor{green!30}\textbf{3.0810e-2 (2.26e-2)} \bigstrut\\
\cline{1-1}\cline{4-11}    MW11  &       &       & \cellcolor{green!30}\textbf{5.0571e-3 (5.65e-4) $+$} & 4.2404e-1 (2.16e-1) $-$ & 4.2692e-1 (2.15e-1) $-$ & 6.8088e-3 (5.57e-4) $+$ & \cellcolor{yellow!30}5.3435e-3 (3.30e-4) $+$ & NaN (NaN) $-$ & 1.1503e-2 (2.35e-3) $+$ & 8.7934e-2 (1.59e-1) \bigstrut\\
\cline{1-1}\cline{4-11}    MW12  &       &       & 3.8856e-2 (1.44e-1) $-$ & 2.0023e-2 (5.37e-2) $-$ & 2.3445e-2 (3.36e-2) $\approx$ & 6.9817e-3 (3.93e-4) $-$ & \cellcolor{yellow!30}6.2835e-3 (2.27e-4) $-$ & 7.2115e-1 (5.87e-2) $-$ & 1.0250e-1 (2.36e-1) $-$ & \cellcolor{green!30}\textbf{5.6167e-3 (4.81e-4)} \bigstrut\\
\cline{1-1}\cline{4-11}    MW13  &       &       & \cellcolor{yellow!30}3.5687e-2 (2.08e-2) $\approx$ & 2.1288e-1 (2.17e-1) $-$ & 1.0502e-1 (7.61e-2) $-$ & 6.0893e-2 (3.57e-2) $-$ & 8.2249e-2 (4.02e-2) $-$ & 5.5828e-1 (1.46e-1) $-$ & 1.9151e-1 (1.02e-1) $-$ & \cellcolor{green!30}\textbf{3.1560e-2 (1.68e-2)} \bigstrut\\
\cline{1-2}\cline{4-11}    MW14  & 3     &       & 9.8945e-2 (1.99e-2) $-$ & 1.2743e-1 (2.81e-3) $-$ & 9.0412e-2 (2.97e-2) $-$ & 9.9136e-2 (5.48e-3) $-$ & \cellcolor{yellow!30}8.6208e-2 (2.74e-3) $-$ & 8.1768e-1 (2.43e-1) $-$ & 1.3570e-1 (4.23e-2) $-$ & \cellcolor{green!30}\textbf{7.7216e-2 (3.25e-3)} \bigstrut\\
    \hline
    \multicolumn{3}{|c|}{$+/-$/$\approx$} & 3/6/5 & 0/11/3 & 2/7/5 & 3/11/0 & 4/10/0 & 0/14/0 & 2/10/2 &  \bigstrut\\
    \hline
\end{tabular}}
  \label{tab:RE1}%
\end{table*}

\begin{table*}[!t]
  \centering
  \caption{Results on CTP test Suite.}
    \resizebox{18cm}{!}{\begin{tabular}{|c|c|c|c|c|c|c|c|c|c|c|}
    \hline
    Problem & \textit{m} & \textit{D} & CCMO  & $\theta$-DEA-CPBI & ShiP  & DRLOS-EMCMO & DVCEA & CMOEA-EoH & CMOEA-ParEvo & CMOEA-STABLE \bigstrut\\
    \hline
    CTP1  & \multirow{7}[14]{*}{2} & \multirow{7}[14]{*}{10} & 1.3234e+0 (1.11e+0) $-$ & 1.1322e+0 (7.60e-1) $-$ & \cellcolor{yellow!30}4.6480e-1 (6.46e-1) $-$ & 4.9270e-1 (5.71e-1) $\approx$ & 4.8827e-1 (5.10e-1) $\approx$ & 2.8199e+0 (1.25e+0) $-$ & 2.0424e+0 (1.05e+0) $-$ & \cellcolor{green!30}\textbf{1.5121e-1 (2.99e-1)} \bigstrut\\
\cline{1-1}\cline{4-11}    CTP2  &       &       & 1.2772e+0 (8.18e-1) $-$ & 1.0449e+0 (7.37e-1) $-$ & 6.3803e-1 (8.60e-1) $-$ & \cellcolor{yellow!30}4.6365e-1 (7.15e-1) $\approx$ & 5.0461e-1 (6.50e-1) $-$ & 2.6592e+0 (1.06e+0) $-$ & 1.7927e+0 (1.37e+0) $-$ & \cellcolor{green!30}\textbf{8.5590e-2 (1.68e-1)} \bigstrut\\
\cline{1-1}\cline{4-11}    CTP3  &       &       & 9.6157e-1 (7.50e-1) $-$ & 8.4067e-1 (9.10e-1) $-$ & 5.5914e-1 (5.24e-1) $-$ & \cellcolor{yellow!30}3.4098e-1 (6.70e-1) $\approx$ & 6.6789e-1 (9.34e-1) $-$ & 3.6457e+0 (1.30e+0) $-$ & 1.6343e+0 (1.17e+0) $-$ & \cellcolor{green!30}\textbf{1.0415e-1 (2.01e-1)} \bigstrut\\
\cline{1-1}\cline{4-11}    CTP4  &       &       & 1.3941e+0 (8.33e-1) $-$ & 9.9763e-1 (8.28e-1) $-$ & 4.5680e-1 (2.88e-1) $-$ & \cellcolor{yellow!30}3.8154e-1 (4.31e-1) $-$ & 5.9768e-1 (5.82e-1) $-$ & 3.4480e+0 (9.73e-1) $-$ & 1.5009e+0 (1.27e+0) $-$ & \cellcolor{green!30}\textbf{2.2291e-1 (3.32e-1)} \bigstrut\\
\cline{1-1}\cline{4-11}    CTP5  &       &       & 1.0707e+0 (7.61e-1) $-$ & 6.3818e-1 (7.73e-1) $-$ & 3.7256e-1 (5.08e-1) $\approx$ & \cellcolor{yellow!30}9.2206e-2 (1.84e-1) $-$ & 8.3322e-1 (8.02e-1) $-$ & 2.4369e+0 (1.27e+0) $-$ & 1.3386e+0 (9.80e-1) $-$ & \cellcolor{green!30}\textbf{7.0625e-2 (2.87e-1)} \bigstrut\\
\cline{1-1}\cline{4-11}    CTP6  &       &       & 5.4922e-1 (7.09e-1) $-$ & 2.3983e+0 (2.27e+0) $-$ & 9.1839e-1 (1.71e+0) $-$ & \cellcolor{green!30}\textbf{1.2497e-2 (8.33e-4) $\approx$} & 7.3430e-1 (1.08e+0) $-$ & 1.1846e+0 (7.53e-1) $-$ & 2.0982e+0 (2.01e+0) $-$ & \cellcolor{yellow!30}2.4963e-2 (5.45e-2) \bigstrut\\
\cline{1-1}\cline{4-11}    CTP7  &       &       & 1.2879e+0 (1.09e+0) $-$ & 1.4904e+0 (1.15e+0) $-$ & 8.2366e-1 (7.36e-1) $-$ & \cellcolor{yellow!30}3.4562e-1 (5.92e-1) $-$ & 2.2601e+0 (1.26e+0) $-$ & 3.1299e+0 (1.17e+0) $-$ & 1.8680e+0 (1.22e+0) $-$ & \cellcolor{green!30}\textbf{8.5375e-2 (2.59e-1)} \bigstrut\\
    \hline
    \multicolumn{3}{|c|}{$+/-$/$\approx$} & 0/7/0 & 0/7/0 & 0/6/1 & 0/3/4 & 0/6/1 & 0/7/0 & 0/7/0 &  \bigstrut\\
    \hline
\end{tabular}}
  \label{tab:RE2}%
\end{table*}

\begin{table*}[!t]
  \centering
  \caption{Results on DTLZ test Suite.}
    \resizebox{18cm}{!}{\begin{tabular}{|c|c|c|c|c|c|c|c|c|c|c|}
    \hline
    Problem & \textit{m} & \textit{D} & K-RVEA & MCEA/D & EDN-ARMOEA & SSDE  & DISK  & SAMOEA-EoH & SAMOEA-ParEvo & SAMOEA-STABLE \bigstrut\\
    \hline
    DTLZ1 & \multirow{7}[14]{*}{3} & \multirow{7}[14]{*}{10} & 8.8355e+1 (1.74e+1) $-$ & \cellcolor{yellow!30}7.2068e+1 (1.93e+1) $\approx$ & 1.1376e+2 (1.87e+1) $-$ & 9.0821e+1 (1.85e+1) $-$ & \cellcolor{green!30}\textbf{7.0244e+1 (2.39e+1) $\approx$} & 7.7630e+1 (1.42e+1) $\approx$ & 1.0146e+2 (1.91e+1) $-$ & 7.4838e+1 (2.27e+1) \bigstrut\\
\cline{1-1}\cline{4-11}    DTLZ2 &       &       & 2.1706e-1 (5.25e-2) $-$ & 2.2795e-1 (2.84e-2) $-$ & 3.1590e-1 (3.48e-2) $-$ & 2.9989e-1 (4.83e-2) $-$ & \cellcolor{green!30}\textbf{5.6436e-2 (7.02e-3) $+$} & 3.6478e-1 (7.68e-2) $-$ & 7.6128e-2 (1.47e-2) $\approx$ & \cellcolor{yellow!30}7.3927e-2 (8.59e-3) \bigstrut\\
\cline{1-1}\cline{4-11}    DTLZ3 &       &       & 2.6991e+2 (4.42e+1) $-$ & \cellcolor{green!30}\textbf{1.5873e+2 (3.78e+1) $+$} & 3.2777e+2 (5.62e+1) $-$ & 2.2831e+2 (4.27e+1) $-$ & 2.0974e+2 (4.66e+1) $\approx$ & \cellcolor{yellow!30}1.7512e+2 (1.96e+1) $+$ & 3.0505e+2 (5.17e+1) $-$ & 1.9355e+2 (3.76e+1) \bigstrut\\
\cline{1-1}\cline{4-11}    DTLZ4 &       &       & \cellcolor{green!30}\textbf{3.1453e-1 (8.47e-2) $+$} & 4.5779e-1 (7.69e-2) $\approx$ & 4.1377e-1 (6.24e-2) $+$ & \cellcolor{yellow!30}3.8438e-1 (7.98e-2) $+$ & 4.3790e-1 (1.33e-1) $\approx$ & 5.7609e-1 (1.00e-1) $-$ & 4.0978e-1 (1.28e-1) $\approx$ & 4.7631e-1 (1.42e-1) \bigstrut\\
\cline{1-1}\cline{4-11}    DTLZ5 &       &       & 1.5711e-1 (3.66e-2) $-$ & 8.4432e-2 (1.89e-2) $-$ & 2.1826e-1 (2.97e-2) $-$ & 2.0519e-1 (3.36e-2) $-$ & \cellcolor{yellow!30}1.9965e-2 (1.03e-2) $\approx$ & 2.4595e-1 (5.85e-2) $-$ & 5.2510e-2 (6.88e-3) $-$ & \cellcolor{green!30}\textbf{1.9525e-2 (3.90e-3)} \bigstrut\\
\cline{1-1}\cline{4-11}    DTLZ6 &       &       & 3.5373e+0 (4.92e-1) $-$ & 2.9432e+0 (6.59e-1) $-$ & 6.2777e+0 (2.94e-1) $-$ & 4.0219e+0 (5.86e-1) $-$ & 3.1216e+0 (8.50e-1) $-$ & \cellcolor{green!30}\textbf{9.4572e-1 (6.99e-1) $+$} & 5.7141e+0 (5.69e-1) $-$ & \cellcolor{yellow!30}2.0097e+0 (5.25e-1) \bigstrut\\
\cline{1-1}\cline{4-11}    DTLZ7 &       &       & \cellcolor{yellow!30}2.4091e-1 (1.95e-1) $-$ & 2.5800e+0 (1.82e+0) $-$ & 3.2387e+0 (7.95e-1) $-$ & 4.6766e+0 (1.75e+0) $-$ & 8.0322e-1 (9.67e-1) $-$ & 2.7091e-1 (9.36e-2) $-$ & 5.1425e+0 (1.06e+0) $-$ &\cellcolor{green!30} \textbf{7.6384e-2 (4.09e-2)} \bigstrut\\
    \hline
    \multicolumn{3}{|c|}{$+/-$/$\approx$} & 1/6/0 & 1/4/2 & 1/6/0 & 1/6/0 & 1/2/4 & 2/4/1 & 0/5/2 &  \bigstrut\\
    \hline
\end{tabular}}
  \label{tab:RE3}%
\end{table*}

\begin{table*}[!t]
  \centering
  \caption{Results on UF test Suite.}
    \resizebox{18cm}{!}{\begin{tabular}{|c|c|c|c|c|c|c|c|c|c|c|}
    \hline
    Problem & \textit{m} & \textit{D} & K-RVEA & MCEA/D & EDN-ARMOEA & SSDE  & DISK  & SAMOEA-EoH & SAMOEA-ParEvo & SAMOEA-STABLE \bigstrut\\
    \hline
    UF1   & \multirow{7}[14]{*}{2} & \multirow{10}[20]{*}{10} & \cellcolor{yellow!30}1.9465e-1 (6.35e-2) $\approx$ & 5.9997e-1 (1.32e-1) $-$ & 6.0964e-1 (1.23e-1) $-$ & 5.9599e-1 (1.64e-1) $-$ & 2.8996e-1 (6.70e-2) $-$ & 5.8647e-1 (1.84e-1) $-$ & 7.3229e-1 (1.98e-1) $-$ & \cellcolor{green!30}\textbf{1.9067e-1 (6.23e-2)} \bigstrut\\
\cline{1-1}\cline{4-11}    UF2   &       &       & \cellcolor{yellow!30}1.2544e-1 (3.16e-2) $-$ & 1.6445e-1 (3.13e-2) $-$ & 1.6634e-1 (3.02e-2) $-$ & 2.9111e-1 (6.89e-2) $-$ & 2.4210e-1 (6.27e-2) $-$ & 1.4246e-1 (2.00e-2) $-$ & 1.4489e-1 (3.17e-2) $-$ & \cellcolor{green!30}\textbf{9.9790e-2 (2.59e-2)} \bigstrut\\
\cline{1-1}\cline{4-11}    UF3   &       &       & \cellcolor{yellow!30}9.0344e-1 (1.66e-1) $-$ & 1.0007e+0 (1.48e-1) $-$ & 1.1661e+0 (1.24e-1) $-$ & 1.2939e+0 (2.14e-1) $-$ & 1.4050e+0 (1.92e-1) $-$ & 9.4452e-1 (1.35e-1) $-$ & 1.3547e+0 (2.80e-1) $-$ & \cellcolor{green!30}\textbf{7.3390e-1 (2.24e-1)} \bigstrut\\
\cline{1-1}\cline{4-11}    UF4   &       &       & \cellcolor{yellow!30}1.1455e-1 (1.38e-2) $\approx$ & 1.3030e-1 (9.16e-3) $-$ & 1.3790e-1 (9.75e-3) $-$ & 1.4054e-1 (7.27e-3) $-$ & 1.4578e-1 (7.60e-3) $-$ & 1.3692e-1 (9.63e-3) $-$ & 1.2405e-1 (1.38e-2) $-$ & \cellcolor{green!30}\textbf{1.0665e-1 (1.40e-2)} \bigstrut\\
\cline{1-1}\cline{4-11}    UF5   &       &       & \cellcolor{green!30}\textbf{1.7683e+0 (7.04e-1) $+$} & 2.9105e+0 (6.12e-1) $\approx$ & 3.5506e+0 (6.24e-1) $-$ & 3.3009e+0 (6.90e-1) $-$ & 2.9384e+0 (4.85e-1) $\approx$ & 2.9061e+0 (6.54e-1) $\approx$ & 3.2577e+0 (7.07e-1) $-$ & \cellcolor{yellow!30}2.7529e+0 (6.59e-1) \bigstrut\\
\cline{1-1}\cline{4-11}    UF6   &       &       & \cellcolor{green!30}\textbf{1.3455e+0 (3.10e-1) $+$} & 2.8382e+0 (9.35e-1) $-$ & 3.4661e+0 (9.13e-1) $-$ & 3.2706e+0 (6.73e-1) $-$ & 2.6265e+0 (6.37e-1) $-$ & 3.2589e+0 (8.50e-1) $-$ & 3.9849e+0 (9.69e-1) $-$ & \cellcolor{yellow!30}2.0066e+0 (5.40e-1) \bigstrut\\
\cline{1-1}\cline{4-11}    UF7   &       &       & \cellcolor{green!30}\textbf{2.6749e-1 (9.77e-2) $\approx$} & 6.2471e-1 (1.90e-1) $-$ & 5.8208e-1 (1.30e-1) $-$ & 6.3160e-1 (1.19e-1) $-$ & 4.4568e-1 (1.03e-1) $-$ & 6.2232e-1 (1.47e-1) $-$ & 6.7834e-1 (2.21e-1) $-$ & \cellcolor{yellow!30}3.2789e-1 (1.05e-1) \bigstrut\\
\cline{1-2}\cline{4-11}    UF8   & \multirow{3}[6]{*}{3} &       & \cellcolor{yellow!30}3.4989e-1 (4.18e-2) $-$ & 6.2764e-1 (1.42e-1) $-$ & 8.8423e-1 (2.21e-1) $-$ & 1.1710e+0 (4.61e-1) $-$ & 3.6683e-1 (7.68e-2) $-$ & 4.6921e-1 (3.99e-2) $-$ & 3.5751e-1 (1.01e-1) $-$ & \cellcolor{green!30}\textbf{2.2629e-1 (2.28e-2)} \bigstrut\\
\cline{1-1}\cline{4-11}    UF9   &       &       & 4.1673e-1 (7.87e-2) $-$ & 6.1126e-1 (1.17e-1) $-$ & 1.1182e+0 (4.06e-1) $-$ & 1.1251e+0 (3.00e-1) $-$ & \cellcolor{yellow!30}3.6765e-1 (6.01e-2) $-$ & 5.2602e-1 (6.82e-2) $-$ & 6.4569e-1 (5.29e-2) $-$ & \cellcolor{green!30}\textbf{2.8801e-1 (3.91e-2)} \bigstrut\\
\cline{1-1}\cline{4-11}    UF10  &       &       & 2.9935e+0 (4.10e-1) $\approx$ & 3.9675e+0 (9.71e-1) $-$ & 5.1857e+0 (9.98e-1) $-$ & 6.7471e+0 (1.56e+0) $-$ & 4.6323e+0 (1.18e+0) $-$ & \cellcolor{yellow!30}2.7238e+0 (4.24e-1) $+$ & \cellcolor{green!30}\textbf{1.5982e+0 (9.10e-1) $+$} & 3.1793e+0 (5.75e-1) \bigstrut\\
    \hline
    \multicolumn{3}{|c|}{$+/-$/$\approx$} & 2/4/4 & 0/9/1 & 0/10/0 & 0/10/0 & 0/9/1 & 1/8/1 & 1/9/0 &  \bigstrut\\
    \hline
\end{tabular}}
  \label{tab:RE4}%
\end{table*}

\subsubsection{Performance of CMOEA-STABLE}
From Table \ref{tab:RE1}, on MW test suite, CMOEA-STABLE is the best algorithm on eight test problems (i.e., MW1, MW3, MW6, MW7, MW10, and MW12--MW14) and performs the second best on MW2 and MW8. For the remaining test problems, CCMO shows promising performance on MW11, Ship ranks the best on MW2 and MW8, DRLOS-EMCMO outperforms others on MW5 and MW9, and CMOEA-ParEvo exhibits the best result on MW4. In addition, although DVCEA fails to achieve the best performance on any test problem, it obtains the second-best results on half of test problems. According to the Wilcoxon's rank-sum test, CMOEA-STABLE is superior to CCMO, $\theta$-DEA-CPBI, Ship, DRLOS-EMCMO, DVCEA, CMOEA-EoH, and CMOEA-ParEvo on six, 11, seven, 11, 10, 14, and 10 test problems, and is inferior to the seven competitors on at most four test problems.

As shown in Table \ref{tab:RE2}, on CTP test suite, CMOEA-STABLE obtains the best $\textup{IGD}^{+}$ values on all seven test problems except CTP6. On CTP6, CMOEA-STABLE yields the second-best result, and DRLOS-EMCMO performs the best. In addition, Ship is the second-best algorithm on CTP1, and DRLOS-EMCMO provides the second-best results on the rest of test problems. Notably, the $\textup{IGD}^{+}$ values achieved by CMOEA-STABLE are even one order of magnitude lower than those of the compared algorithms. The Wilcoxon's rank-sum test results show that CMOEA-STABLE beats CCMO, $\theta$-DEA-CPBI, Ship, DRLOS-EMCMO, DVCEA, CMOEA-EoH, and CMOEA-ParEvo on seven, seven, six, three, six, seven, and seven test problems, respectively. However, the seven peer algorithms cannot outperform CMOEA-STABLE on any test problems.

\subsubsection{Performance of SAMOEA-STABLE}
We can observe from Table \ref{tab:RE3}, on DTLZ test suite, K-RVEA, MCEA/D, EDN-ARMOEA, SSDE,  DISK, SAMOEA-EoH, SAMOEA-ParEvo, and SAMOEA-STABLE are the best algorithms on one, one, zero, zero, two, one, zero, and two test problems, respectively, and the second-best algorithms on one, one, zero, one, one, one, zero, and two test problems, respectively. From the Wilcoxon's rank-sum test results, SAMOEA-STABLE is better than K-RVEA, MCEA/D, EDN-ARMOEA, SSDE,  DISK, SAMOEA-EoH, and SAMOEA-ParEvo on six, four, six, six, two, four, and five test problems, respectively, and is worse than them on one, one, one, one, one, two, and zero test problems, respectively.

As listed in Table \ref{tab:RE4}, on UF test suite, SAMOEA-STABLE and K-RVEA are the two best-performing algorithms. To be specifical, SAMOEA-STABLE performs the best on six out of ten test problems and the second best on three test problems, while K-RVEA ranks the best on three test problems and the second best on five test problems. As for the other algorithms, DISK and SAMOEA-EoH each record the second-best performance on only one test problem, and SAMOEA-ParEvo yields the best $\textup{IGD}^{+}$ value on one test problem. In terms of the Wilcoxon's rank-sum test, SAMOEA-STABLE surpasses K-RVEA, MCEA/D, EDN-ARMOEA, SSDE,  DISK, SAMOEA-EoH, and SAMOEA-ParEvo on four, nine, ten, ten, nine, eight, and nine test problems, respectively, and is beaten by them on no more than two test problems.

Based on the above experimental results, both CMOEA-STABLE and SAMOEA-STABLE outperform not only state-of-the-art hand-crafted algorithms, but also the top-performing approaches produced by advanced LES methods. This firmly verifies the effectiveness of STABLE as a powerful LES method for AMAD.

\section{Conclusion}\label{Sec:Con}
This paper came up with a customized LES method for AMAD, called STABLE (Semantics-Aware Bilevel Co-Evolution), a unified framework that integrates structured algorithm formulation and semantics-driven evolutionary search. STABLE abandons the traditional flat representation and models complex algorithms as hierarchical, modular architectures based on domain expertise, which aligns the search space with the intrinsic compositional characteristics of complex evolutionary algorithms. Building on this structured formulation, STABLE adopts a bilevel co-evolutionary mechanism: the upper level explores the full-algorithm space to generate diverse and structurally coherent complete algorithms, while the lower level performs fine-grained search in each component space to optimize individual algorithm components. The two levels collaborate synergistically, balancing global exploration and local exploitation to avoid both coarse-grained inefficiency and fragmented component optimization. Additionally, STABLE incorporates a semantics-aware composite genetic operator, which first clusters parent individuals based on code structural similarity and then performs probabilistic intra-cluster or inter-cluster crossover, as well as targeted mutation, to generate semantically consistent and functionally improved offspring individuals based on five-dimensional semantics modeling. To efficiently allocate limited computational resources, a semantics-aware performance evaluator is further proposed, which adaptively selects between true evaluation and fitness inheritance based on a control probability related to code structural similarity and current APE consumption. This integrated design enables STABLE to significantly improve the search efficiency and performance of AMAD. The experiments on two AMAD tasks validated the superiority of STABLE over advanced LES methods, as well as the competitive performance of the top STABLE-generated algorithms compared with advanced human-designed counterparts.

\bibliographystyle{IEEEtran}

\bibliography{myref}
\end{document}